\pdfoutput=1
\documentclass[11pt]{article}

\usepackage{acl}
\usepackage{booktabs}
\usepackage{hyperref}
\usepackage{url}
\usepackage{pifont}
\usepackage{graphicx}
\usepackage{tabularx}
\usepackage{soul, color}
\usepackage{amsmath}
\usepackage[most]{tcolorbox}
\usepackage{caption}
\usepackage{wrapfig}
\usepackage{amssymb}
\usepackage[table,svgnames,dvipsnames]{xcolor}
\definecolor{CustomOrange}{HTML}{ffd98d} 
\definecolor{CustomGreen}{HTML}{abe0b6}
\definecolor{lightgreen}{HTML}{D5F5E3}
\definecolor{lightred}{HTML}{FADBD8}

\usepackage[table]{xcolor}
\usepackage{makecell,multirow}

\usepackage{times}
\usepackage{latexsym}
\usepackage[T1]{fontenc}
\usepackage[utf8]{inputenc}

\usepackage{microtype}

\usepackage{inconsolata}

\usepackage{enumitem}

\title{MPR-GUI: Benchmarking and Enhancing Multilingual \\ Perception and Reasoning in GUI Agents}

\author{
  Ruihan Chen$^{1}$\thanks{~Equal Contribution}, 
  Qiming Li$^{1\ast}$\thanks{~Project Lead.}, 
  Xiaocheng Feng$^{1,2}$\thanks{~Crresponding Author.}, 
  Weihong Zhong$^1$, \\ \textbf{Xiaoliang Yang$^1$, Yuxuan Gu$^1$, Zekun Zhou$^1$, Yunfei Lu$^3$,Haoyu Ren$^3$,}\\\textbf{Kun Chen$^3$,Dandan Tu$^3$, Bing Qin$^{1,2}$}
   \\
  $^1$Harbin Institute of Technology \quad $^2$Peng Cheng Laboratory \\$^3$Huawei Technologies Co., Ltd \\
  \texttt{\{rhchen, qmli\}@ir.hit.edu.cn} \\
  \vspace{1ex} \\
}

\begin{document}
\maketitle

\begin{abstract}
Large Vision–Language Models (LVLMs) have shown strong potential as multilingual Graphical User Interface (GUI) agents, as evidenced by existing GUI benchmarks. However, these benchmarks exhibit two primary limitations: (1) although Perception and Reasoning (P\&R) capabilities are fundamental for GUI agents, current benchmarks lack fine-grained diagnostics to identify which specific capabilities lead to task failures, hindering targeted improvements; (2) existing benchmarks fail to provide a strictly aligned cross-lingual evaluation environment, introducing confounding factors that prevent isolating the language impact on GUI agent performance. To address these issues, we propose the \textbf{M}ultilingual \textbf{P\&R} \textbf{GUI} \textbf{Bench}mark (\textbf{MPR-GUI-Bench}), featuring strictly aligned environments across six languages and eight fine-grained P\&R tasks. Our benchmark reveals consistent P\&R gaps between English and non-English settings, particularly on reasoning-intensive tasks. To leverage the superior English P\&R capabilities for bridging cross-lingual gaps, we identify layers sensitive to language and propose \textbf{GUI-XLI}, a \textbf{GUI Cross-Lingual I}ntervention method that aligns non-English hidden states with their English counterparts at these layers during inference. Experiments show that GUI-XLI effectively reduces the cross-lingual gaps, with an average gain of 6.5\% in non-English settings.
\end{abstract}

\section{Introduction}
\begin{figure}[t]
    \centering
    \includegraphics[width=1.0\columnwidth]{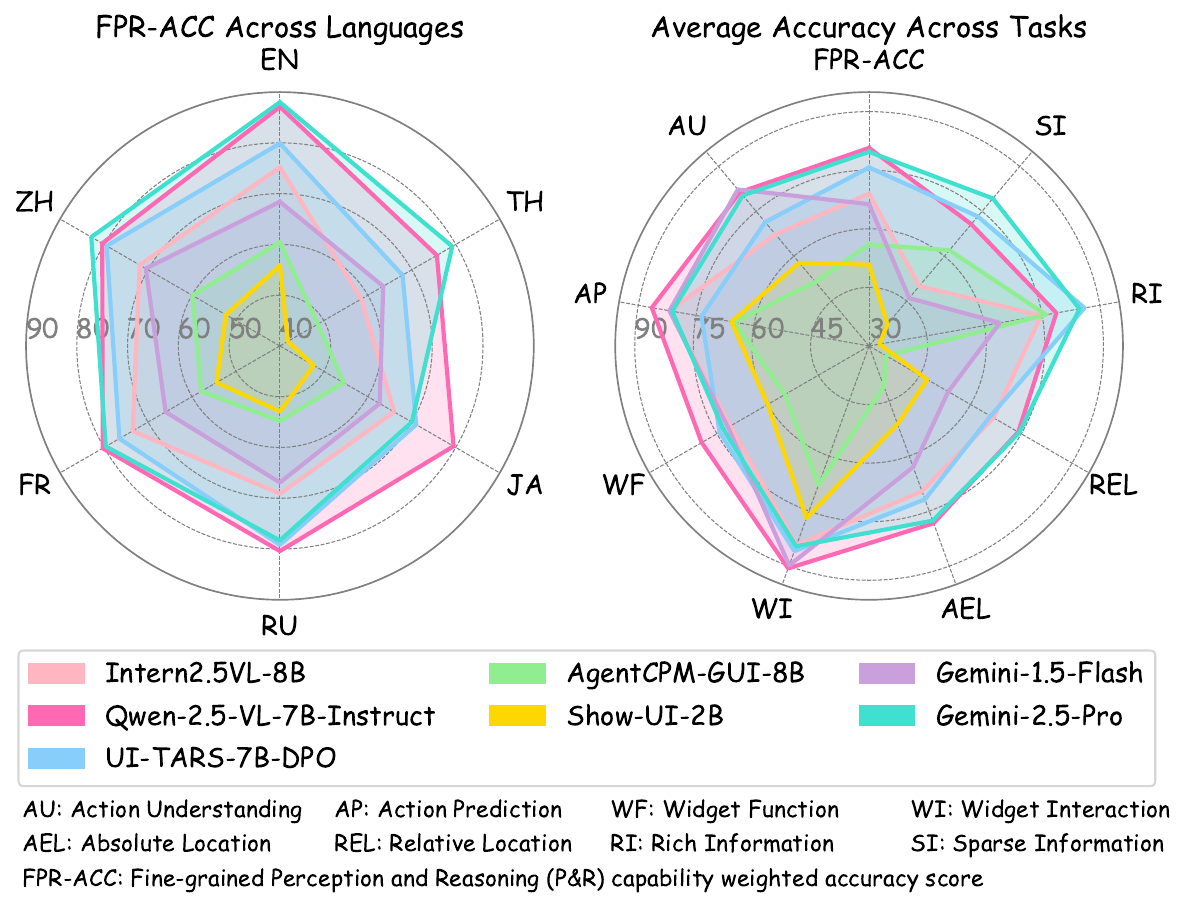}
 \caption{\textbf{Performance evaluation on MPR-GUI-Bench.} \textbf{(Left)} Multilingual comparison revealing a consistent gap between English and non-English settings across all baselines. \textbf{(Right)} Fine-grained P\&R capability analysis across specific dimensions.}
    \label{fig:radar}
\end{figure}

\begin{table*}[!t]
\centering
\scriptsize
\setlength{\tabcolsep}{4.2pt} 
\begin{tabular}{l|cccccc|c|c|c|ccc|cc}
\toprule[1.5pt]
\multirow{2}{*}{\textbf{Dataset}} & \multicolumn{6}{c|}{\textbf{Languages}} & \multirow{2}{*}{\textbf{\begin{tabular}[c]{@{}c@{}}Lang.\\ Type\end{tabular}}} & \multirow{2}{*}{\textbf{\begin{tabular}[c]{@{}c@{}}Cross-lingual\\ alignment\end{tabular}}} & \multirow{2}{*}{\textbf{\begin{tabular}[c]{@{}c@{}}Fine-grained \\Dimensions\end{tabular}}} & \multicolumn{3}{c|}{\textbf{Platform}} & \multirow{2}{*}{\textbf{Size}} & \multirow{2}{*}{\textbf{Type}} \\
& EN & ZH & FR & RU & JA & TI & & & & Web. & Mob. & Desk. & & \\
\midrule[1pt]
GUI-WORLD & \cellcolor{CustomGreen!60}\ding{52} & \ding{55} & \ding{55} & \ding{55} & \ding{55} & \ding{55} & 1 & \ding{55} & \ding{55} & \ding{52} & \ding{52} & \ding{52} & 12,379 & \textit{dataset} \\
\midrule
OSWorld & \cellcolor{CustomGreen!60}\ding{52} & \ding{55} & \ding{55} & \ding{55} & \ding{55} & \ding{55} & 1 & \ding{55} & \ding{55} & \ding{52} & \ding{52} & \ding{52} & 369 & \textit{env.} \\
\midrule
AndroidWorld & \cellcolor{CustomGreen!60}\ding{52} & \ding{55} & \ding{55} & \ding{55} & \ding{55} & \ding{55} & 1 & \ding{55} & \ding{55} & \ding{55} & \ding{52} & \ding{55} & 116 & \textit{env.} \\
\midrule
Mobile-Agent-Bench & \cellcolor{CustomGreen!60}\ding{52} & \ding{55} & \ding{55} & \ding{55} & \ding{55} & \ding{55} & 1 & \ding{55} & \ding{55} & \ding{55} & \ding{52} & \ding{55} & 100 & \textit{env.} \\
\midrule
ScreenSpot & \cellcolor{CustomGreen!60}\ding{52} & \ding{55} & \ding{55} & \ding{55} & \ding{55} & \ding{55} & 1 & \ding{55} & \ding{55} & \ding{52} & \ding{52} & \ding{52} & 1200+ & \textit{dataset} \\
\midrule
GUI-Odyssey & \cellcolor{CustomGreen!60}\ding{52} & \ding{55} & \ding{55} & \ding{55} & \ding{55} & \ding{55} & 1 & \ding{55} & \ding{55} & \ding{55} & \ding{52} & \ding{55} & 7735 & \textit{dataset} \\
\midrule
CAGUI & \ding{55} & \cellcolor{CustomGreen!60}\ding{52} & \ding{55} & \ding{55} & \ding{55} & \ding{55} & 1 & \ding{55} & \ding{55} & \ding{55} & \ding{52} & \ding{55} & $\sim$1K-10K & \textit{dataset} \\
\midrule
SPA-Bench & \cellcolor{CustomGreen!60}\ding{52} & \cellcolor{CustomGreen!60}\ding{52} & \ding{55} & \ding{55} & \ding{55} & \ding{55} & 2 & \ding{55} & \ding{55} & \ding{55} & \ding{52} & \ding{55} & 340 & \textit{env.} \\
\midrule
WebMMU & \cellcolor{CustomGreen!60}\ding{52} & \ding{55} & \cellcolor{CustomGreen!60}\ding{52} & \ding{55} & \ding{55} & \ding{55} & 4 & \ding{55} & \ding{55} & \ding{52} & \ding{55} & \ding{55} & $\sim$3K+ & \textit{dataset} \\
\midrule
MacOSWorld & \cellcolor{CustomGreen!60}\ding{52} & \cellcolor{CustomGreen!60}\ding{52} & \cellcolor{CustomGreen!60}\ding{52} & \cellcolor{CustomGreen!60}\ding{52} & \ding{55} & \ding{55} & 5 & \ding{55} & \ding{55} & \ding{55} & \ding{55} & \ding{52} & 201+29 & \textit{env.} \\
\midrule
X-WebAgentBench & \cellcolor{CustomGreen!60}\ding{52} & \cellcolor{CustomGreen!60}\ding{52} & \cellcolor{CustomGreen!60}\ding{52} & \cellcolor{CustomGreen!60}\ding{52} & \ding{55} & \ding{55} & 14 & \ding{55} & \cellcolor{CustomGreen!60}\ding{52} & \ding{52} & \ding{55} & \ding{55} & 2800 & \textit{env.} \\
\midrule
\textbf{MPR-GUI-Bench} & \cellcolor{CustomGreen!60}\ding{52} & \cellcolor{CustomGreen!60}\ding{52} & \cellcolor{CustomGreen!60}\ding{52} & \cellcolor{CustomGreen!60}\ding{52} & \cellcolor{CustomGreen!60}\ding{52} & \cellcolor{CustomGreen!60}\ding{52} & 6 & \cellcolor{CustomGreen!60}\ding{52} & \cellcolor{CustomGreen!60}\ding{52} & \ding{52} & \ding{52} & \ding{55} & 12,936 & \textit{dataset} \\
\bottomrule[1.5pt]
\end{tabular}
\caption{Comparison with prevailing benchmarks. Abbreviations for \textbf{Platform} include \textbf{Web.} (Website), \textbf{Mob.} (Mobile), and \textbf{Desk.} (Desktop). For the \textbf{Type} column, \textit{env.} denotes interactive environments, while \textit{dataset} refers to static data collections. \textbf{Cross-lingual alignment} indicates whether the benchmark provides strictly aligned tasks across different languages to facilitate controlled cross-lingual evaluation.}
\label{tab:benchmark_comparison}
\end{table*}
Rapidly evolving Large Vision-Language Models (LVLMs) have shown potential as multilingual GUI agents, as demonstrated by recent benchmarks. Existing GUI benchmarks, broadly categorized into interactive and static ones, suffer from two critical limitations. First, despite GUI Perception and Reasoning (P\&R) capabilities being fundamental to real-world end-to-end competence~\citep{zhang-etal-2025-agentcpm,xie-etal-2025-gui,qin2025ui}, current benchmarks lack systematic and fine-grained assessment of these capabilities. Interactive benchmarks rely on holistic task success rates, obscuring failure causes, while static benchmarks lack structured P\&R analysis. Second, existing benchmarks lack strictly aligned cross-lingual evaluation. Linguistic factors are underexplored in static benchmarks, while interactive benchmarks such as MacOSWorld~\citep{macosworld} inevitably introduce language-irrelevant variations (e.g., UI layouts and interaction trajectories), preventing isolation of language effects.

To bridge these gaps, we introduce the \textbf{M}ultilingual \textbf{P\&R GUI Bench}mark (\textbf{MPR-GUI-Bench}). MPR-GUI-Bench is strictly aligned across six languages carefully chosen to balance common GUI languages and orthogonal coverage of different language families, spanning 39 scenarios and six device types with 12,936 samples. Building upon key perception and reasoning tasks proposed by prior works, we further incorporate end-to reasoning tasks that reflect real-world end-to-end scenarios, resulting in tasks hierarchically organized into three levels and eight fine-grained P\&R dimensions. As shown in Figure~\ref{fig:radar}, evaluations across seven advanced LVLMs reveal consistent non-English performance gaps relative to English, particularly in reasoning tasks.

To leverage the superior P\&R capabilities of English for improving non-English performance, we identify critical layers most sensitive to linguistic factors during inference and propose a \textbf{GUI Cross-L}ingual \textbf{I}ntervention method (\textbf{GUI-XLI}), which steers non-English representations toward their English counterparts. GUI-XLI achieves a 6.5\% average performance gain with negligible inference latency. Consistent improvements under explicit reasoning-chain generation indicate that GUI-XLI aligns cross-lingual reasoning patterns at the representational level rather than acting as a prompting artifact.

Our contributions are summarized as follows: \begin{itemize}[leftmargin=10pt] \item We present \textbf{MPR-GUI-Bench}, the first multilingual benchmark to systematically evaluate fine-grained GUI P\&R capabilities.
\item We present a comprehensive analysis of the LVLMs for GUI agent from the perspectives of P\&R and cross-lingual capabilities.
\item We propose \textbf{GUI-XLI}, a training-free representation engineering method that effectively mitigates cross-lingual P\&R capability gaps. 
\end{itemize}

\section{Related Work}
\paragraph{GUI agent benchmarks}
As presented in Table~\ref{tab:benchmark_comparison}, existing GUI agent benchmarks generally fall into two categories: interactive environments and static datasets~\citep{nguyen2024guiagentssurvey}. Interactive environments are widely regarded as better reflections of real-world end-to-end scenarios~\citep{rawles2024androidworlddynamicbenchmarkingenvironment,wang2024mobileagentbenchefficientuserfriendlybenchmark,chen2025spabench,xwebagentbench,OSWorld}. However, they typically evaluate performance through task completion rates, treating each trajectory as a single unit. Such a coarse-grained metric obscures the underlying P\&R skills required for success, making it difficult to diagnose failures or guide targeted improvements. Conversely, static datasets~\citep{zhang-etal-2025-agentcpm,webmmu} such as ScreenSpot~\citep{cheng2024seeclick}, GUI-World~\citep{chen2024gui} and GUI-Odessey~\citep{lu2024guiodyssey} isolate particular capabilities and demonstrate the value of decomposing GUI tasks into foundational skills, but they lack a systematic and fine-grained taxonomy. Moreover, multilingual capability, a prerequisite for global deployment, remains largely under-explored. While interactive benchmarks like MacOSWorld~\citep{macosworld} include language factors, their dynamic nature introduces complexity, hindering controlled, aligned comparison across languages. These limitations highlight the need for a systematic, multilingual, fine-grained P\&R evaluation setup, which motivates the design of MPR-GUI-Bench.
\begin{figure}[t]
    \centering
    \includegraphics[width=0.9\columnwidth]{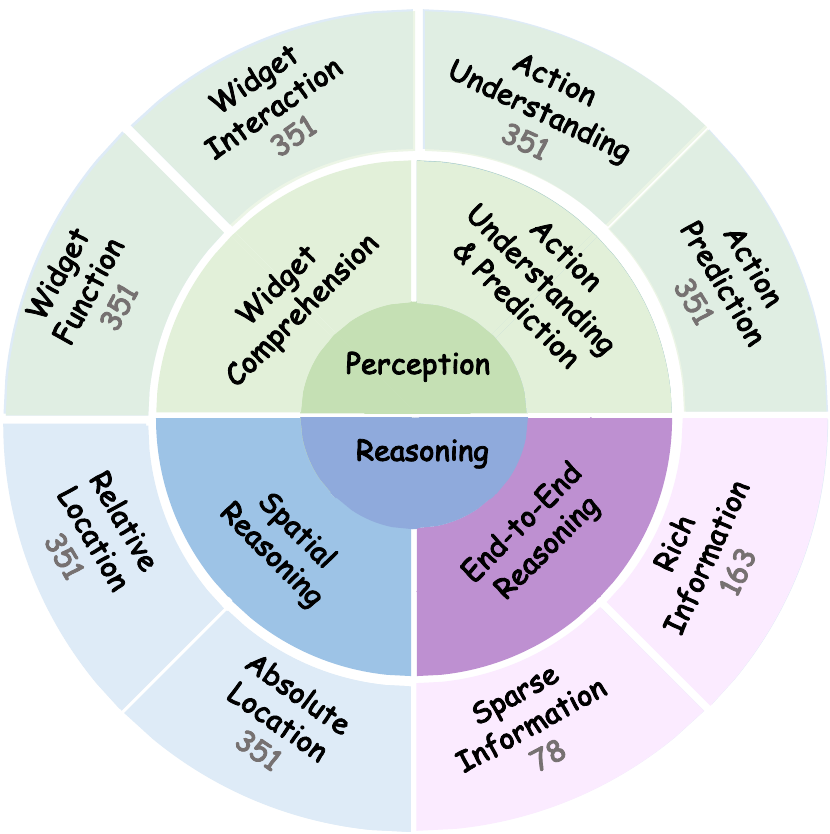}
     \caption{Hierarchical composition of MPR-GUI-Bench. The structure expands from P\&R (inner) to four domains (middle) and eight fine-grained dimensions (outer). Gray numbers indicate image counts.}
    \label{fig:composition}
\end{figure}
\begin{figure*}[t]
    \centering
    \includegraphics[width=1.0\textwidth]{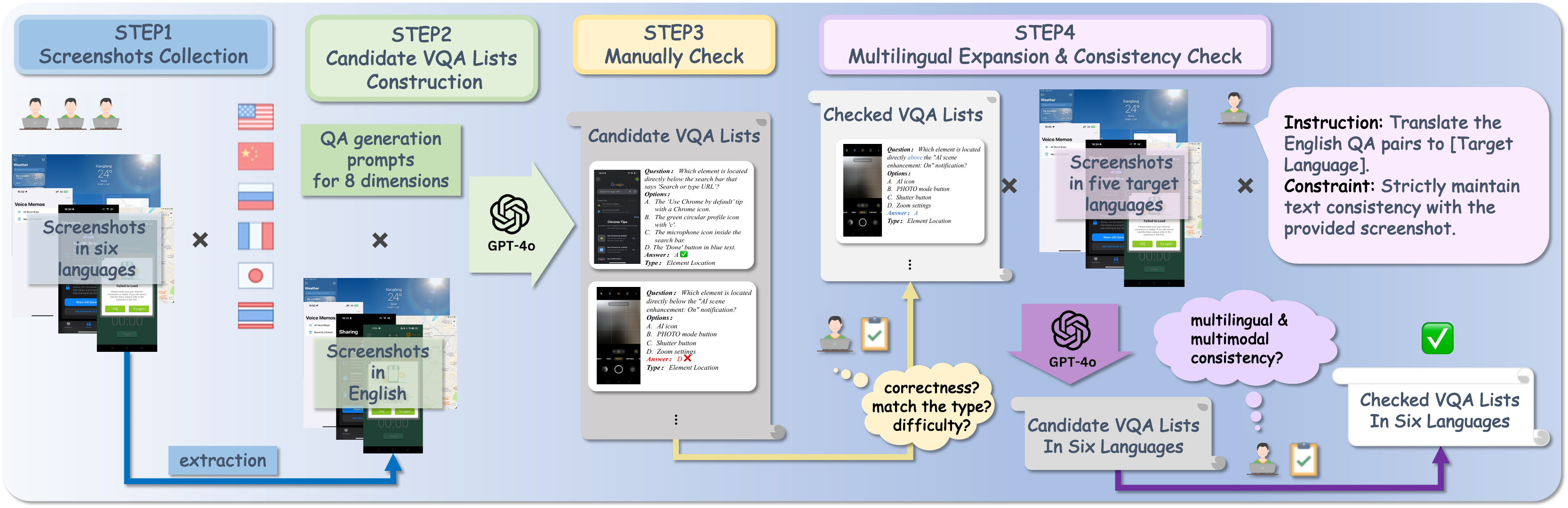}
    \caption{Construction pipeline of MPR-GUI-Bench, as described in \S\ref{Benchmark Construction Pipeline}: 
\textbf{Step 1}: Collect parallel screenshots in six languages; 
\textbf{Step 2}: Generate candidate English VQA lists via GPT-4o; 
\textbf{Step 3}: Verify quality manually; 
\textbf{Step 4}: Expand multilingual data with GPT-4o and perform cross-lingual and cross-modal consistency checks.}
    \label{fig2}
\end{figure*}

\paragraph{Cross-lingual Representation Alignment}
Recent studies in LVLMs have investigated the internal mechanisms of cross-lingual transfer. Research indicates that inputs with identical semantics but different languages often yield substantially different distributions within the model's latent space~\citep{chang-etal-2022-geometry,peng2025debiasingmultilingualllmscrosslingual,LLMhandleMultilingualism}. This divergence correlates with performance inconsistencies across languages. Meanwhile, representation engineering research demonstrates that controlled interventions on hidden states can effectively steer model behavior~\citep{zou2023transparency,li2024inference, turner2024steeringlanguagemodelsactivation,li2025unlockingmultilingualreasoningcapability}. Notably, recent work further explores training-free strategies to enhance the semantic comprehension of LVLMs without additional fine-tuning~\citep{xiao2026not,QIN2025101118,qin-etal-2023-cross,xiao2026not}. Inspired by these findings, our proposed GUI-XLI treats the cross-lingual discrepancy vector as an explicit optimization direction, aligning under-performing languages toward the distribution of better-performing ones, thereby enhancing multilingual GUI P\&R consistency.



\section{MPR-GUI-Bench}
\label{gen_inst}

Existing GUI benchmarks have mostly neglected fine-grained P\&R capabilities, leading to difficulties in their development. Moreover, even fewer studies have focused on these capabilities in multilingual settings. To this end, We propose \textbf{MPR-GUI-Bench}, the first benchmark to systematically evaluate the fine-grained P\&R capabilities required by GUI tasks in multilingual environments.

\subsection{Data Source}
As shown in Figure \ref{fig2}, we collect parallel screenshots across 6 languages (English, Chinese, French, Russian, Japanese and Thai), spanning 39 distinct real-world GUI scenarios on two operating systems (iOS and Android) and 6 mobile device models. 

\subsection{Task Definitions}
\label{definitions}
As shown in Figure \ref{fig:composition}, we define 8 fine-grained dimensions derived from key P\&R capabilities mentioned in prior works, supplemented by two dimensions reflecting end-to-end agent performance, organized into two primary categories: (1) \textbf{perception capabilities}, covering the perception of interactive components (widgets) and user actions; and (2) \textbf{reasoning capabilities}, which encompass spatial reasoning and end-to-end reasoning based on integrating fundamental P\&R capabilities.
The eight dimensions are defined as follows: 

\paragraph{Perception Capabilities Evaluation Dimensions}
\begin{itemize}[leftmargin=10pt,itemsep=1pt,parsep=0pt, topsep=3pt]
    \item \textbf{Widget Function Comprehension (WF)} evaluates LVLMs' perception of the function of GUI elements and the meaning of visual cues.
    \item \textbf{Widget Interaction Comprehension (WI)} evaluates LVLMs' perception of the most suitable way for users to interaction with widgets.
    \item \textbf{Action Understanding (AU)} evaluates LVLMs' perception of the consequences of executed actions, including interface changes, system feedback, and impacts on future interactions.
    \item \textbf{Action Prediction (AP)} evaluates LVLMs' perception of action organization (e.g., types, targets, order, input content) to accomplish goals.
\end{itemize}

\paragraph{Reasoning Capabilities Evaluation Dimensions}
\begin{itemize}[leftmargin=10pt,itemsep=1pt,parsep=0pt, topsep=3pt]
\item \textbf{Absolute Element Location (AEL)} evaluates LVLMs' reasoning capability to correctly locate UI elements and analyze their global positions.
\item \textbf{Relative Element Location (REL)} evaluates LVLMs' reasoning capability in relative spatial relationships between GUI elements.

\item \textbf{Rich Information (RI)} evaluates the capabilities to synthesize long interaction histories, and integrate all fine-grained P\&R dimensions to infer user intention. It simulates real-world interactive end-to-end GUI tasks. 
\item \textbf{Sparse Information (SI)} evaluates the capability to infer user intent from shorter screenshot sequences with minimal cues. It simulates more challenging real-world end-to-end GUI tasks, reflecting the model's P\&R capability upper bound in more complex GUI tasks.
\end{itemize}

\subsection{Benchmark Construction Pipeline}
\label{Benchmark Construction Pipeline}
As illustrated in Figure~\ref{fig2}, to balance scalability with accuracy, we adopt a semi-automated approach followed by human verification:

\noindent\textbf{Step 1: Screenshot Collection}
Annotators curated a rigorously aligned dataset across six languages. While end-to-end reasoning tasks feature fewer samples due to the complexity of aligning multi-image sequences, they offer deeper diagnostic insights than standard single-image tasks. Detailed information about the annotators' backgrounds and guidelines are provided in the Appendix~\ref{appendix:Annotator Background},~\ref{appendix:Data Collection Guidelines}.

\noindent\textbf{Step 2: Candidate Visual Question Answering(VQA) Lists Construction} Following the definitions in Section~\ref{definitions}, we construct prompts that enforce strict structural, lexical constraints, limiting the model’s generative freedom 
and reducing the imprinting of model-specific phrasing style. GPT-4o~\citep{openai2024gpt4ocard} is then used to generate the English VQA candidates based on the prompts, which are provided in Appendix~\ref{apdx:Prompts For Candidate VQA Lists Construction}.
\label{Step 2}

\newcolumntype{C}[1]{>{\centering\arraybackslash}p{#1}}
\newcolumntype{L}[1]{>{\raggedright\arraybackslash}p{#1}}
\begin{table*}[!ht]
\centering
\scriptsize
\renewcommand{\arraystretch}{0.9} 
\setlength{\tabcolsep}{5pt} 

\begin{tabular}{ L{3.2cm} C{0.6cm} C{0.85cm} C{0.85cm} C{0.85cm} C{0.85cm} C{0.85cm} C{0.85cm} C{0.85cm} C{0.85cm} C{1.25cm} } 
\toprule 
\multirow{2}{*}{\textbf{Model}} & \multirow{2}{*}{\textbf{Lang}} & \multicolumn{4}{c}{\textbf{Perception}} & \multicolumn{4}{c}{\textbf{Reasoning}} & \multirow{2}{*}{\textbf{$\text{FPR-ACC}$}} \\ 
\cmidrule(lr){3-6} \cmidrule(lr){7-10}
 &  & \textbf{AU} & \textbf{AP} & \textbf{WF} & \textbf{WI} & \textbf{AEL} & \textbf{REL} & \textbf{RI} & \textbf{SI} &  \\ 
\midrule 

\rowcolor[gray]{0.95} \multicolumn{11}{c}{\textit{Open-source LVLMs}} \\

\multirow{6}{*}{Intern2.5VL-8B} & EN & \cellcolor{CustomGreen!60}\textbf{81.2} & \cellcolor{CustomGreen!60}\textbf{89.9} & \cellcolor{CustomGreen!60}\textbf{79.5} & \cellcolor{CustomGreen!60}\textbf{92.1} & \cellcolor{CustomGreen!60}\textbf{82.0} & \cellcolor{CustomGreen!60}\textbf{82.0} & \cellcolor{CustomGreen!60}\textbf{80.0} & \cellcolor{CustomOrange!60}44.0 & \cellcolor{CustomGreen!60}\textbf{75.2} \\ 
& ZH & 72.4 & 85.5 & 75.1 & 88.0 & 78.4 & 67.8 & \cellcolor{CustomOrange!60}64.0 & \cellcolor{CustomGreen!60}\textbf{60.0} & 71.9 \\ 
& FR & 77.1 & 83.9 & 75.6 & 88.5 & 72.7 & 76.5 & \cellcolor{CustomGreen!60}\textbf{80.0} & 52.0 & 73.5 \\ 
& RU & 70.2 & 81.4 & 70.4 & 83.3 & 68.3 & 66.9 & \cellcolor{CustomGreen!60}\textbf{80.0} & 48.0 & 69.1 \\ 
& JA & 64.2 & 82.8 & 72.9 & 80.6 & 73.2 & 69.1 & \cellcolor{CustomOrange!60}64.0 & \cellcolor{CustomOrange!60}44.0 & 66.0 \\ 
& TH & \cellcolor{CustomOrange!60}57.9 & \cellcolor{CustomOrange!60}67.5 & \cellcolor{CustomOrange!60}52.6 & \cellcolor{CustomOrange!60}72.7 & \cellcolor{CustomOrange!60}42.9 & \cellcolor{CustomOrange!60}38.3 & \cellcolor{CustomGreen!60}\textbf{80.0} & 52.0 & \cellcolor{CustomOrange!60}58.5 \\ 
\midrule 

\multirow{6}{*}{Qwen-2.5-VL-7B-Instruct} & EN & \cellcolor{CustomGreen!60}\textbf{86.1} & \cellcolor{CustomGreen!60}\textbf{89.4} & \cellcolor{CustomGreen!60}\textbf{86.0} & \cellcolor{CustomGreen!60}\textbf{93.4} & \cellcolor{CustomGreen!60}\textbf{86.0} & \cellcolor{CustomGreen!60}\textbf{81.6} & \cellcolor{CustomGreen!60}\textbf{96.0} & \cellcolor{CustomGreen!60}\textbf{72.0} & \cellcolor{CustomGreen!60}\textbf{87.1} \\ 
& ZH & 83.6 & 88.8 & 77.8 & \cellcolor{CustomOrange!60}88.8 & 79.2 & 74.3 & \cellcolor{CustomOrange!60}68.0 & \cellcolor{CustomOrange!60}68.0 & 80.4 \\ 
& FR & 81.7 & 83.6 & 80.0 & 91.3 & 76.5 & 79.0 & 72.0 & \cellcolor{CustomGreen!60}\textbf{72.0} & 80.3 \\ 
& RU & 77.6 & 86.1 & 76.7 & 89.6 & 77.3 & 75.1 & 72.0 & \cellcolor{CustomGreen!60}\textbf{72.0} & 80.4 \\ 
& JA & 81.7 & 87.7 & 79.2 & 90.7 & 77.3 & 69.1 & 88.0 & \cellcolor{CustomOrange!60}68.0 & 79.5 \\ 
& TH & \cellcolor{CustomOrange!60}76.8 & \cellcolor{CustomOrange!60}82.5 & \cellcolor{CustomOrange!60}77.5 & \cellcolor{CustomOrange!60}88.8 & \cellcolor{CustomOrange!60}73.5 & \cellcolor{CustomOrange!60}65.3 & 76.0 & \cellcolor{CustomGreen!60}\textbf{72.0} & \cellcolor{CustomOrange!60}75.7 \\ 
\midrule 

\rowcolor[gray]{0.95} \multicolumn{11}{c}{\textit{Multimodal GUI Agents}} \\

\multirow{6}{*}{UI-TARS-7B-DPO} & EN & 76.2 & \cellcolor{CustomGreen!60}\textbf{79.5} & \cellcolor{CustomGreen!60}\textbf{80.8} & \cellcolor{CustomGreen!60}\textbf{88.8} & 74.0 & 71.3 & \cellcolor{CustomOrange!60}80.0 & \cellcolor{CustomGreen!60}\textbf{84.0} & \cellcolor{CustomGreen!60}\textbf{79.9} \\ 
& ZH & \cellcolor{CustomGreen!60}\textbf{77.9} & 74.9 & 72.6 & 88.5 & \cellcolor{CustomGreen!60}\textbf{77.9} & 66.1 & \cellcolor{CustomGreen!60}\textbf{96.0} & 76.0 & 79.4 \\ 
& FR & 67.5 & 70.8 & 77.0 & 85.5 & 69.7 & 70.5 & 84.0 & 80.0 & 76.5 \\ 
& RU & 72.1 & 75.1 & 73.7 & 87.4 & 71.0 & \cellcolor{CustomGreen!60}\textbf{72.7} & 88.0 & \cellcolor{CustomGreen!60}\textbf{84.0} & 79.2 \\ 
& JA & \cellcolor{CustomOrange!60}67.2 & 73.5 & 72.3 & 85.8 & 70.5 & 63.1 & 84.0 & \cellcolor{CustomOrange!60}58.0 & 71.0 \\ 
& TH & \cellcolor{CustomOrange!60}67.2 & \cellcolor{CustomOrange!60}67.5 & \cellcolor{CustomOrange!60}69.0 & \cellcolor{CustomOrange!60}79.2 & \cellcolor{CustomOrange!60}67.8 & \cellcolor{CustomOrange!60}51.9 & 84.0 & \cellcolor{CustomOrange!60}58.0 & \cellcolor{CustomOrange!60}67.9 \\ 
\midrule 

\multirow{6}{*}{AgentCPM-GUI-8B} & EN & \cellcolor{CustomGreen!60}\textbf{64.8} & \cellcolor{CustomGreen!60}\textbf{71.6} & \cellcolor{CustomGreen!60}\textbf{67.1} & \cellcolor{CustomGreen!60}\textbf{82.8} & \cellcolor{CustomGreen!60}\textbf{57.1} & \cellcolor{CustomGreen!60}\textbf{39.3} & 80.0 & 72.0 & \cellcolor{CustomGreen!60}\textbf{60.4} \\ 
& ZH & 49.2 & 64.8 & 58.1 & 70.5 & 41.8 & 32.8 & \cellcolor{CustomGreen!60}\textbf{88.0} & \cellcolor{CustomGreen!60}\textbf{88.0} & 59.9 \\ 
& FR & 54.6 & 64.5 & 57.3 & 69.1 & 40.0 & 36.1 & 72.0 & 60.0 & 57.9 \\ 
& RU & \cellcolor{CustomOrange!60}43.4 & 62.3 & 53.7 & 62.6 & 46.0 & 32.2 & 72.0 & 56.0 & 54.8 \\ 
& JA & 51.4 & 68.6 & 50.1 & 66.4 & 33.8 & 37.4 & 76.0 & \cellcolor{CustomOrange!60}48.0 & 54.5 \\ 
& TH & 45.6 & \cellcolor{CustomOrange!60}59.3 & \cellcolor{CustomOrange!60}42.7 & \cellcolor{CustomOrange!60}56.0 & \cellcolor{CustomOrange!60}27.9 & \cellcolor{CustomOrange!60}32.0 & \cellcolor{CustomOrange!60}68.0 & \cellcolor{CustomOrange!60}48.0 & \cellcolor{CustomOrange!60}48.6 \\ 
\midrule 

\multirow{6}{*}{Show-UI-2B} & EN & \cellcolor{CustomGreen!60}\textbf{67.8} & \cellcolor{CustomGreen!60}\textbf{73.8} & \cellcolor{CustomGreen!60}\textbf{64.7} & \cellcolor{CustomGreen!60}\textbf{84.7} & 57.1 & \cellcolor{CustomGreen!60}\textbf{62.3} & \cellcolor{CustomOrange!60}32.0 & \cellcolor{CustomOrange!60}36.0 & \cellcolor{CustomGreen!60}\textbf{55.8} \\ 
& ZH & 62.8 & 66.1 & 61.4 & 82.2 & \cellcolor{CustomGreen!60}\textbf{58.5} & 45.6 & \cellcolor{CustomOrange!60}32.0 & \cellcolor{CustomOrange!60}36.0 & 52.3 \\ 
& FR & 59.0 & 67.2 & 61.6 & 79.0 & 55.5 & 48.4 & \cellcolor{CustomGreen!60}\textbf{44.0} & \cellcolor{CustomGreen!60}\textbf{40.0} & 54.4 \\ 
& RU & 55.7 & 68.9 & 61.6 & 77.6 & 50.3 & 48.1 & 40.0 & \cellcolor{CustomGreen!60}\textbf{40.0} & 52.9 \\ 
& JA & 54.6 & 64.2 & 57.3 & 75.4 & 44.5 & 42.6 & 28.0 & \cellcolor{CustomOrange!60}36.0 & 47.7 \\ 
& TH & \cellcolor{CustomOrange!60}46.7 & \cellcolor{CustomOrange!60}54.6 & \cellcolor{CustomOrange!60}53.2 & \cellcolor{CustomOrange!60}62.6 & \cellcolor{CustomOrange!60}41.5 & \cellcolor{CustomOrange!60}36.1 & \cellcolor{CustomOrange!60}20.0 & \cellcolor{CustomOrange!60}36.0 & \cellcolor{CustomOrange!60}41.8 \\ 
\midrule 

\rowcolor[gray]{0.95} \multicolumn{11}{c}{\textit{Close-source LVLMs}} \\

\multirow{6}{*}{Gemini-1.5-Flash} & EN & 85.0 & \cellcolor{CustomGreen!60}\textbf{85.8} & \cellcolor{CustomGreen!60}\textbf{76.2} & \cellcolor{CustomGreen!60}\textbf{93.4} & \cellcolor{CustomGreen!60}\textbf{71.6} & 61.5 & 64.0 & 40.0 & 68.4 \\ 
& ZH & \cellcolor{CustomGreen!60}\textbf{86.2} & 81.4 & 68.5 & 89.9 & 64.5 & 49.2 & \cellcolor{CustomGreen!60}\textbf{68.0} & \cellcolor{CustomGreen!60}\textbf{64.0} & \cellcolor{CustomGreen!60}\textbf{70.5} \\ 
& FR & 84.4 & 80.6 & 74.0 & 90.4 & 64.8 & \cellcolor{CustomGreen!60}\textbf{65.0} & 64.0 & \cellcolor{CustomOrange!60}36.0 & 66.0 \\ 
& RU & 80.1 & 81.2 & 72.6 & 89.9 & 66.1 & 59.3 & \cellcolor{CustomOrange!60}60.0 & 48.0 & 66.9 \\ 
& JA & 80.3 & 82.8 & 71.2 & 88.8 & \cellcolor{CustomOrange!60}52.0 & 44.7 & 64.0 & 40.0 & \cellcolor{CustomOrange!60}62.7 \\ 
& TH & \cellcolor{CustomOrange!60}77.9 & \cellcolor{CustomOrange!60}79.5 & \cellcolor{CustomOrange!60}67.7 & \cellcolor{CustomOrange!60}86.3 & 59.0 & \cellcolor{CustomOrange!60}40.4 & 64.0 & 48.0 & 63.5 \\
\midrule 

\multirow{6}{*}{Gemini-2.5-Pro} & EN & 85.0 & \cellcolor{CustomGreen!60}\textbf{90.7} & \cellcolor{CustomGreen!60}\textbf{85.0} & \cellcolor{CustomGreen!60}\textbf{93.2} & \cellcolor{CustomGreen!60}\textbf{84.7} & \cellcolor{CustomGreen!60}\textbf{93.2} & \cellcolor{CustomGreen!60}\textbf{96.0} & 80.0 & \cellcolor{CustomGreen!60}\textbf{88.0} \\
& ZH & 78.4 & 85.8 & 82.5 & 81.2 & 82.5 & 71.0 & 92.0 & \cellcolor{CustomGreen!60}\textbf{84.0} & 82.9 \\
& FR & \cellcolor{CustomGreen!60}\textbf{86.9} & \cellcolor{CustomOrange!60}64.5 & 81.6 & 92.9 & 81.4 & 65.3 & 88.0 & \cellcolor{CustomOrange!60}76.0 & 79.6 \\
& RU & \cellcolor{CustomOrange!60}63.4 & 90.4 & 54.0 & 92.1 & 75.4 & 70.8 & 92.0 & 80.0 & 78.3 \\
& JA & 85.2 & 84.7 & \cellcolor{CustomOrange!60}53.4 & \cellcolor{CustomOrange!60}65.0 & \cellcolor{CustomOrange!60}60.1 & \cellcolor{CustomOrange!60}62.3 & \cellcolor{CustomOrange!60}68.0 & \cellcolor{CustomOrange!60}76.0 & \cellcolor{CustomOrange!60}70.0 \\
& TH & 82.8 & 71.3 & 81.6 & 83.4 & 81.6 & 83.7 & 72.0 & 80.0 & 79.2 \\
\bottomrule
\end{tabular} 
\caption{Model Performance (\%) Across six Languages English, Chinese, France, Russian, Japanese and Thai (EN, ZH, FR, RU, JA, TH). Background colors indicate the \colorbox{CustomGreen!60}{\textbf{best}} and \colorbox{CustomOrange!60}{worst} performance per setting.}
\small
\label{tab:performance_result} 
\end{table*}
\noindent\textbf{Step 3: Manually Check}
Six annotators independently verify the candidate VQA list across three dimensions: question formulation, answer correctness, and distractor quality. To mitigate GPT-4o's stylistic artifacts, annotators manually reorder options and rephrase question patterns. We further validate the removal of source-specific bias by comparing VQA items generated by Gemini-2.5-Pro and GPT-4o. Evaluation with three models reveals no consistent performance discrepancies between the two sources, suggesting the benchmark is robust to generator choice. Detailed guidelines, verification results, and inter-rater agreement statistics are provided in the Appendix~\ref{appendix:Comprehensive Inter-Rater Reliability Analysis for All Languages}.
\label{Step 3}

\noindent\textbf{Step 4: Multilingual Expansion \& Consistency Check}
We employ GPT-4o for translation. Unlike text-only machine translation systems, which often generate synonyms causing lexical mismatches and subsequent GUI grounding failures, GPT-4o's multimodal capabilities ensure the translated text aligns faithfully with the on-screen text.  Analyses of translation fidelity and inter-rater consistency are provided in the Appendix~\ref{appendix:Validation on GPT-4o Translation},~\ref{appendix:Comprehensive Inter-Rater Reliability Analysis for All Languages}. 
\label{Step 4}

\subsection{Evaluation Metrics}
Samples follow a standard four-option single-choice format, evaluated via exact match following~\citep{chen2024gui}. We eschew free-form responses because real-world GUI tasks inherently require precise understanding and selection within a constrained action space; free-form outputs often introduce unnecessary evaluation noise. To represent the overall performance considering dimension difficulty, we define a  fine-grained P\&R weighted accuracy score $\text{FPR-ACC}$:
\begin{equation*}
    \text{FPR-ACC} = \frac{\sum_{i=1}^{8} w_i \cdot \alpha_i}{\sum_{i=1}^{8} w_i},
\end{equation*}
where $\alpha_i$ and $w_i$ denote the accuracy and weight for dimension $i$, respectively. Detailed descriptions are provided in Appendix~\ref{appendix:Details about FPR-ACC}.

\subsection{Experiment Setup}

\paragraph{Baseline}We select baselines from three model types: (1) \textbf{Open-source LVLMs}: Intern2.5VL-8B~\citep{chen2024internvl}, Qwen-2.5-VL-7B-Instruct~\citep{Qwen2.5-VL}; (2) \textbf{Closed-source LVLMs}: Gemini-1.5-flash
~\citep{geminiteam2024gemini15unlockingmultimodal}
and Gemini-2.5-Pro
~\citep{comanici2025gemini25pushingfrontier}; (3) \textbf{Multimodal GUI agents}: UI-TARS-7B-DPO~\citep{qin2025ui}, AgentCPM-GUI~\citep{zhang2025agentcpmgui}, Show-UI-2B~\citep{lin2024showui}.
Among models with known parameter sizes, we select versions smaller than 8B, as lightweight models better support on-device deployment for GUI agents, which is critical for preserving user privacy. The evaluation results for three extra models are provided in Appendix~\ref{appendix:Evaluations results on additional models with broader size}.

\paragraph{Implementation Details} Our evaluation is conducted on 8 × NVIDIA A100 GPUs. 
\subsection{Evaluation Result}
\label{eval-result}
From the evaluation result presented in Table~\ref{tab:performance_result}, we draw conclusions in three key aspects:

\noindent\textbf{Performance gap across languages \& models } 
Regarding languages, while English and Chinese yield dominant results across all baselines, performance degrades in low-resource settings (e.g., Thai), revealing a severe generalization bottleneck in current LVLMs for GUI tasks.
\begin{table}[t]
\centering
\small
\begin{tabular}{lcc}
\toprule
\textbf{Metric} & \textbf{Pearson $r$} & \textbf{Strength} \\ \midrule
RI vs. FPR-ACC  & 0.7373       & High              \\
SI vs. FPR-ACC  & 0.7152       & High              \\
\textbf{Avg. vs. FPR-ACC} & \textbf{0.7795} & \textbf{V. High} \\ \bottomrule
\end{tabular}
\caption{Correlation Analysis of End-to-End Reasoning Scores and FPR-ACC; Avg. denotes the mean of RI and SI scores.\textbf{V. High}: Very High.}
\label{tab:correlation}
\end{table}

\noindent\textbf{Performance gap across dimensions} A significant capability imbalance exists across the eight dimensions. Most models achieve near-saturation in basic perception tasks (e.g., WI). However, performance diverges sharply in spatial reasoning tasks. 

\noindent\textbf{Correlation Between Fundamental P\&R Capabilities and End-to-End Competence.} As shown in Table~\ref{tab:correlation}, RI and SI—two tasks with varying difficulty levels that reflect end-to-end agent performance—show a high correlation with FPR-ACC. This indicates that  FPR-ACC effectively reflects both the models' fundamental P\&R capabilities and their advanced end-to-end performance.
\section{GUI-XL-Intervention}
Building on prior findings that representation alignment mitigates cross-lingual discrepancies~\citep{NEURIPS2024_a6678e2b,peng2025debiasingmultilingualllmscrosslingual,chang-etal-2022-geometry}, we aim to leverage the superior P\&R capabilities of English to bridge cross-lingual P\&R gaps. To this end, we identify the layers where the cross-lingual input representation distributions are most divergent, as exemplified in the left graph of Figure~\ref{fig:tsne}. We then propose \textbf{GUI Cross-Lingual Intervention (GUI-XLI)} to steer non-English representations toward their English counterparts (See Appendix~\ref{appendix:Overview of GUI-XLI} for an overview).

\subsection{Preliminaries}
Given an LVLM parameterized by $\theta$, visual and textual inputs are embedded and concatenated into an initial sequence $\boldsymbol{H}^{(0)}$. Processed through $L$ Transformer~\citep{transformers} layers, the hidden state $h^{(l)}$ of the final token updates via:
\begin{equation}
h^{(l)} = h^{(l-1)} + a^{(l)} + m^{(l)},
\end{equation}
where $a^{(l)}$ and $m^{(l)}$ denote attention and MLP outputs, respectively. Finally, $\boldsymbol{H}^{(L)}$ is projected by the language head for autoregressive prediction.
\subsection{Cross-lingual Discrepancy Vector Construction}
\label{Cross-lingual Discrepancy Vector Construction}
For an input question $Q$ and image $I$, we denote the hidden state of the last token at layer $l$ as:
$h^{(l)}(Q, I) \in \mathbb{R}^d$. 
We denote the test input as $X$ when it includes a question, a reasoning chain, and the final answer. Correspondingly, the hidden state of the last token at layer $l$ is denoted as:
$h^{(l)}(X, I) \in \mathbb{R}^d$. 
When the inputs are in English or the target language, we denote them with a subscript $en$ or $tgt$, respectively.

Given the autoregressive nature of LVLMs, the final token's hidden state aggregates the global context of the input sequence. Consequently, it serves as a natural anchor for cross-lingual alignment. Leveraging this property, we construct cross-lingual discrepancy vectors $\delta_{\textbf{en-tgt}}^{(l)}$ between parallel inputs $(X_{\textbf{en}}, I_{\textbf{en}})$ and $(X_{\textbf{tgt}}, I_{\textbf{tgt}})$ to faithfully capture the gaps in GUI P\&R capabilities:
\begin{equation}
\delta_{\textbf{en-tgt}}^{(l)} = h^{(l)}(X_{\textbf{en}}, I_{\textbf{en}}) - h^{(l)}(X_{\textbf{tgt}}, I_{\textbf{tgt}}),
\end{equation}
where the subscript $\textbf{en}$ and $\textbf{tgt}$ means English and the target language, respectively. The effectiveness of $\delta_{\textbf{en-tgt}}^{(l)}$ in isolating linguistically induced P\&R discrepancies relies on three core design principles:

\paragraph{Explicit Reasoning Chains}Text input $X$ includes the question, a reasoning chain, and the final answer. This explicitly externalized reasoning chain ensures that $h^{(l)}(X, I)$ captures the underlying GUI P\&R logic for choosing the final answer.

\paragraph{Positive–negative Sample Pairs} 
Building on the observation that LVLMs use intermediate layers as an English-centric reasoning hub~\citep{LLMhandleMultilingualism}, we select sample pairs where the model succeeds in English ($X_{\text{en}}$) but fails in the target language ($X_{\text{tgt}}$). This strategy contrasts the "successful English reasoning pathway" against the "failed target-language pathway" within these pivotal layers, maximizing the extraction of the capability gap while minimizing irrelevant linguistic noise.
\paragraph{Visual-Linguistic Isolation} $I_{\text{en}}$ and $I_{\text{tgt}}$ represent identical GUI scenarios differing solely in language, which ensures that any visual variance in $\delta_{\text{en-tgt}}^{(l)}$ stems exclusively from visual-linguistic discrepancies, aligning with MPR-GUI-Bench.
\subsection{GUI-XL-Memory}
\label{sec:GUI-XL-Memory}

We construct the GUI Cross-Lingual Memory (GUI-XL-Memory) to store the cross-lingual discrepancy vectors, which enables LVLMs to adaptively retrieve and apply these vectors as optimization directions during inference, facilitating robust generalization to out-of-domain GUI scenarios.

Specifically, we extract the target language query and image representation as the retrieval key:
\begin{equation}
r_{\textbf{tgt}}^{(l)} = h^{(l)}(Q_{\textbf{tgt}}, I_{\textbf{tgt}}),
\end{equation}
and the discrepancy vector as the value:
\begin{equation}
v_{\textbf{en-tgt}}^{(l)} = h^{(l)}(X_{\textbf{en}}, I_{\textbf{en}}) - h^{(l)}(X_{\textbf{tgt}}, I_{\textbf{tgt}}).
\end{equation}
The $(r_{\textbf{tgt}}^{(l)}, v_{\textbf{en-tgt}}^{(l)})$ forms one entry of the memory.


\newcolumntype{L}[1]{>{\raggedright\arraybackslash}m{#1}}  
\newcolumntype{C}[1]{>{\centering\arraybackslash}m{#1}}

\newcommand{\gainbox}[2]{{\setlength{\fboxsep}{1pt}\colorbox{#1}{#2}}}

\begin{table*}[!ht]
\centering
\scriptsize

\gdef\mcjot{1pt}
\setlength{\tabcolsep}{3pt}

\begin{tabular}{ l c c l l l l l l l l l } 
\toprule 
\multirow{2}{*}{\textbf{Model}} & \multirow{2}{*}{\textbf{Lang}} & \multirow{2}{*}{\textbf{GXI}}
& \multicolumn{4}{c}{\textbf{Perception}} 
& \multicolumn{4}{c}{\textbf{Reasoning}} 
& \multirow{2}{*}{\textbf{FPR-ACC}} \\ 
\cmidrule(lr){4-7} \cmidrule(lr){8-11}
 &  &  & \textbf{AU} & \textbf{AP} & \textbf{WF} & \textbf{WI} & \textbf{AEL} & \textbf{REL} & \textbf{RI} & \textbf{SI} &  \\ 
\midrule

\multirow{3}{*}{Intern2.5VL-8B}
    & ZH & \makecell{$\times$ \\ \checkmark} 
         & \makecell[l]{72.4 \\ 81.9 \gainbox{lightgreen}{$\uparrow$9.5}} 
         & \makecell[l]{85.5 \\ 90.2 \gainbox{lightgreen}{$\uparrow$4.7}} 
         & \makecell[l]{75.1 \\ 80.3 \gainbox{lightgreen}{$\uparrow$5.2}}
         & \makecell[l]{88.0 \\ 90.5 \gainbox{lightgreen}{$\uparrow$2.5}}
         & \makecell[l]{78.4 \\ 81.9 \gainbox{lightgreen}{$\uparrow$3.5}} 
         & \makecell[l]{67.8 \\ 70.2 \gainbox{lightgreen}{$\uparrow$2.4}} 
         & \makecell[l]{64.0 \\ 80.0 \gainbox{lightgreen}{$\uparrow$16.0}}
         & \makecell[l]{60.0 \\ 72.0 \gainbox{lightgreen}{$\uparrow$12.0}}
         & \makecell[l]{71.9 \\ 82.8 \gainbox{lightgreen}{$\uparrow$10.9}}\\
\cmidrule(lr){2-12}
    & TH & \makecell{$\times$ \\ \checkmark} 
         & \makecell[l]{57.9 \\ 58.3 \gainbox{lightgreen}{$\uparrow$0.4}} 
         & \makecell[l]{67.5 \\ 69.4 \gainbox{lightgreen}{$\uparrow$1.9}}
         & \makecell[l]{52.6 \\ 55.8 \gainbox{lightgreen}{$\uparrow$3.2}}
         & \makecell[l]{72.7 \\ 71.7 \gainbox{lightred}{$\downarrow$1.0}} 
         & \makecell[l]{42.9 \\ 44.1 \gainbox{lightgreen}{$\uparrow$1.2}}
         & \makecell[l]{38.3 \\ 39.6 \gainbox{lightgreen}{$\uparrow$1.3}}
         & \makecell[l]{80.0 \\ 80.0 \gainbox{lightgreen}{$\uparrow$0.0}}
         & \makecell[l]{52.0 \\ 40.0 \gainbox{lightred}{$\downarrow$12.0}}
         & \makecell[l]{58.5 \\ 62.2 \gainbox{lightgreen}{$\uparrow$3.7}}\\
\cmidrule(lr){2-12}
    & JA & \makecell{$\times$ \\ \checkmark} 
         & \makecell[l]{64.2 \\ 67.5 \gainbox{lightgreen}{$\uparrow$3.3}} 
         & \makecell[l]{82.8 \\ 85.5 \gainbox{lightgreen}{$\uparrow$2.7}}
         & \makecell[l]{72.9 \\ 75.1 \gainbox{lightgreen}{$\uparrow$2.2}}
         & \makecell[l]{80.6 \\ 81.5 \gainbox{lightgreen}{$\uparrow$0.9}}
         & \makecell[l]{73.2 \\ 72.4 \gainbox{lightred}{$\downarrow$0.8}}
         & \makecell[l]{69.1 \\ 68.7 \gainbox{lightred}{$\downarrow$0.4}} 
         & \makecell[l]{64.0 \\ 72.0 \gainbox{lightgreen}{$\uparrow$8.0}}
         & \makecell[l]{44.0 \\ 56.0 \gainbox{lightgreen}{$\uparrow$12.0}}
         & \makecell[l]{69.1 \\ 77.2 \gainbox{lightgreen}{$\uparrow$8.1}}\\
\midrule

\multirow{3}{*}{Qwen-2.5-VL-Instruct}
    & ZH & \makecell{$\times$ \\ \checkmark} 
         & \makecell[l]{83.6 \\ 86.2 \gainbox{lightgreen}{$\uparrow$2.6}} 
         & \makecell[l]{88.8 \\ 89.2 \gainbox{lightgreen}{$\uparrow$0.4}} 
         & \makecell[l]{77.8 \\ 78.1 \gainbox{lightgreen}{$\uparrow$0.3}} 
         & \makecell[l]{88.8 \\ 91.5 \gainbox{lightgreen}{$\uparrow$2.7}} 
         & \makecell[l]{79.2 \\ 79.6 \gainbox{lightgreen}{$\uparrow$0.4}} 
         & \makecell[l]{74.3 \\ 74.6 \gainbox{lightgreen}{$\uparrow$0.3}} 
         & \makecell[l]{68.0 \\ 84.0 \gainbox{lightgreen}{$\uparrow$16.0}}
         & \makecell[l]{68.0 \\ 76.0 \gainbox{lightgreen}{$\uparrow$8.0}}
         & \makecell[l]{77.1 \\ 83.1 \gainbox{lightgreen}{$\uparrow$6.0}}\\
\cmidrule(lr){2-12}
    & RU & \makecell{$\times$ \\ \checkmark} 
         & \makecell[l]{77.6 \\ 84.6 \gainbox{lightgreen}{$\uparrow$7.0}} 
         & \makecell[l]{86.1 \\ 87.9 \gainbox{lightgreen}{$\uparrow$1.8}} 
         & \makecell[l]{76.7 \\ 77.4 \gainbox{lightgreen}{$\uparrow$0.7}} 
         & \makecell[l]{89.6 \\ 88.6 \gainbox{lightred}{$\downarrow$1.0}} 
         & \makecell[l]{77.3 \\ 77.5 \gainbox{lightgreen}{$\uparrow$0.2}} 
         & \makecell[l]{75.1 \\ 74.5 \gainbox{lightred}{$\downarrow$0.6}} 
         & \makecell[l]{72.0 \\ 84.0 \gainbox{lightgreen}{$\uparrow$12.0}}
         & \makecell[l]{72.0 \\ 84.0 \gainbox{lightgreen}{$\uparrow$12.0}}
         & \makecell[l]{78.1 \\ 83.6 \gainbox{lightgreen}{$\uparrow$5.5}}\\
\cmidrule(lr){2-12}
    & JA & \makecell{$\times$ \\ \checkmark} 
         & \makecell[l]{81.7 \\ 83.6 \gainbox{lightgreen}{$\uparrow$1.9}} 
         & \makecell[l]{87.7 \\ 88.3 \gainbox{lightgreen}{$\uparrow$0.6}} 
         & \makecell[l]{79.2 \\ 81.7 \gainbox{lightgreen}{$\uparrow$2.5}} 
         & \makecell[l]{90.2 \\ 93.5 \gainbox{lightgreen}{$\uparrow$3.3}} 
         & \makecell[l]{77.3 \\ 77.1 \gainbox{lightred}{$\downarrow$0.2}} 
         & \makecell[l]{69.1 \\ 69.3 \gainbox{lightgreen}{$\uparrow$0.2}} 
         & \makecell[l]{88.0 \\ 92.0 \gainbox{lightgreen}{$\uparrow$4.0}}
         & \makecell[l]{68.0 \\ 80.0 \gainbox{lightgreen}{$\uparrow$12.0}}
         & \makecell[l]{79.9 \\ 84.4 \gainbox{lightgreen}{$\uparrow$4.5}}\\
\bottomrule
\end{tabular}
\caption{Effectiveness of GUI-XLI across languages. GXI denotes GUI-XLI. Each cell reports accuracy (\%) for the baseline ($\times$) and our method ($\checkmark$). Colored highlights denote absolute performance \colorbox{lightgreen}{gains} or \colorbox{lightred}{drops}.}
\label{tab:method_comparison}
\end{table*}

\subsection{Cross-lingual Representation Intervention}
\label{sec:Intervention}
Given the current input $(Q_{\text{tgt}}^{\text{curr}}, I_{\text{tgt}}^{\text{curr}})$, we extract the hidden state $h_{\text{tgt}}^{(l,\text{curr})}$. We then identify the top-$k$ semantically nearest entries in GUI-XL-Memory by maximizing cosine similarity:
\begin{equation}
\mathcal{I} = \underset{\mathcal{J}\subseteq\{1,\dots,N\},\,|\mathcal{J}|=k}{\operatorname{arg\,max}}
\;\sum_{i\in\mathcal{J}} 
\frac{(h_{\textbf{tgt}}^{(l,\text{curr})})^\top r_{i}^{(l)}}{\|h_{\textbf{tgt}}^{(l,\text{curr})}\|_2 \,\|r_{i}^{(l)}\|_2}.
\end{equation}
The retrieved crossdiscrepancy vectors are averaged to form the intervention vector $\bar{v}_{\text{en-tgt}}^{(l)}$. 
During inference, we intervene on the residual stream by incorporating $\bar{v}_{\text{en-tgt}}^{(l)}$ with strength $\alpha$, followed by magnitude-preserving normalization:
\begin{equation}
\tilde{h}_{\textbf{tgt}}^{(l,\text{curr})} = \frac{\|h_{\textbf{tgt}}^{(l,\text{curr})}\|_2 \cdot \left(h_{\textbf{tgt}}^{(l,\text{curr})} + \alpha \,\bar{v}_{\textbf{en-tgt}}^{(l)}\right)}{\|h_{\textbf{tgt}}^{(l,\text{curr})} + \alpha \,\bar{v}_{\textbf{en-tgt}}^{(l)}\|_2},
\end{equation}
which steers non-English representations toward English P\&R patterns without modifying model parameters. Consequently, GUI-XLI is training-free and compatible with diverse LVLM architectures.
\section{Experiment}
\subsection{Setup}
\paragraph{Baseline Models}We evaluate GUI-XLI's effectiveness on Intern2.5VL-8B~\citep{chen2024internvl} and Qwen2.5-VL-7B-Instruct~\citep{Qwen2.5-VL}.
\paragraph{Language Selection}
We select 4 languages (ZH, JA, RU, TH) as their non-Latin scripts and large linguistic distance from English pose the most challenging cross-lingual alignment conditions.
\paragraph{Memory Construction}
We build dimension-specific memories for each basic P\&R tasks. The entries are randomly selected while ensuring coverage of distinct GUI scenarios. The sampling process follows that of MPR-GUI-Bench, with the only distinction being that we require LVLMs to generate reasoning chains for their answers. To construct memories for end-to-end reasoning tasks (RI and SI), we aggregate entries across the six basic P\&R dimensions, as end-to-end reasoning represents an integrated synthesis of these basic capabilities Details are provided in Appendix~\ref{aappendix:Memory Construction Details}.
\begin{table}[htbp]
\centering
\scriptsize 
\setlength{\tabcolsep}{3pt} 
\begin{tabular}{lcccc}
\toprule
\textbf{Model} & \textbf{Lang.} & \textbf{Layer ($l$)} & \boldmath$\alpha$ & \textbf{Mem. Size} \\
\midrule
\multirow{3}{*}{Qwen-2.5-VL-7B-Instruct} & ZH & 16 & 0.12 & 10 \\
                                  & RU & 18 & 0.1  & 10 \\
                                  & JA & 13 & 0.1  & 10 \\
\midrule
\multirow{3}{*}{Intern2.5VL-8B}   & ZH & 12 & 0.12 & 10 \\
                                  & TH & 17 & 0.06 & 10 \\
                                  & JA & 14 & 0.1  & 10 \\
\bottomrule
\end{tabular}
\caption{Hyper-parameter settings for GUI-XLI across different models and languages.}
\label{tab:hyperparameters}
\end{table}
\paragraph{Hyper-parameter settings} To ensure reproducibility, we provide the hyper-parameter settings for each LVLM and language in Table~\ref{tab:hyperparameters}. Although the optimal parameters vary slightly, all selected layers consistently fall within the intermediate range (layers 11-18), confirming our finding that this is the universal "sweet spot" for cross-lingual alignment. The variations in intervention strength simply reflect the distinct linguistic and typological distances between each target language and English. Furthermore, the optimal settings are efficiently identified through a rapid forward-pass grid search on a minimal validation set, which demonstrates our method's structural adaptability to different architectures and languages without incurring high computational tuning costs.

\begin{table*}[!ht]
\centering
\scriptsize
\gdef\mcjot{1pt}
\setlength{\tabcolsep}{3pt}

\begin{tabular}{ l c c l l l l l l l l l } 
\toprule 
\multirow{2}{*}{\textbf{Model}} & \multirow{2}{*}{\textbf{Lang}} & \multirow{2}{*}{\textbf{GXI}}
& \multicolumn{4}{c}{\textbf{Perception}} 
& \multicolumn{4}{c}{\textbf{Reasoning}} 
& \multirow{2}{*}{\textbf{FPR-ACC}} \\ 
\cmidrule(lr){4-7} \cmidrule(lr){8-11}
 &  &  & \textbf{AU} & \textbf{AP} & \textbf{WF} & \textbf{WI} & \textbf{AEL} & \textbf{REL} & \textbf{RI} & \textbf{SI} &  \\ 
\midrule

\multirow{3}{*}{Qwen-2.5-VL-Instruct}
    & ZH & \makecell{$\times$ \\ \checkmark} 
         & \makecell[l]{77.8 \\ 82.1 \gainbox{lightgreen}{$\uparrow$4.3}} 
         & \makecell[l]{72.9 \\ 83.5 \gainbox{lightgreen}{$\uparrow$10.5}} 
         & \makecell[l]{69.8 \\ 74.4 \gainbox{lightgreen}{$\uparrow$4.6}} 
         & \makecell[l]{67.8 \\ 67.7 \gainbox{lightred}{$\downarrow$0.1}} 
         & \makecell[l]{73.7 \\ 78.3 \gainbox{lightgreen}{$\uparrow$4.6}} 
         & \makecell[l]{80.6 \\ 85.5 \gainbox{lightgreen}{$\uparrow$4.8}} 
         & \makecell[l]{84.0 \\ 80.0 \gainbox{lightgreen}{$\uparrow$0.3}} 
         & \makecell[l]{68.0 \\ 72.0 \gainbox{lightgreen}{$\uparrow$4.0}} 
         & \makecell[l]{74.2 \\ 77.4 \gainbox{lightgreen}{$\uparrow$3.2}} \\
\cmidrule(lr){2-12}
    & RU & \makecell{$\times$ \\ \checkmark} 
         & \makecell[l]{83.5 \\ 88.9 \gainbox{lightgreen}{$\uparrow$5.4}} 
         & \makecell[l]{87.9 \\ 90.7 \gainbox{lightgreen}{$\uparrow$2.8}} 
         & \makecell[l]{76.9 \\ 77.5 \gainbox{lightgreen}{$\uparrow$0.6}} 
         & \makecell[l]{68.4 \\ 68.4 \gainbox{lightgreen}{$\uparrow$0.0}} 
         & \makecell[l]{77.4 \\ 76.4 \gainbox{lightred}{$\downarrow$1.0}} 
         & \makecell[l]{87.7 \\ 87.5 \gainbox{lightred}{$\downarrow$0.2}} 
         & \makecell[l]{84.0 \\ 88.0 \gainbox{lightgreen}{$\uparrow$4.0}} 
         & \makecell[l]{80.0 \\ 80.0 \gainbox{lightgreen}{$\uparrow$0.0}} 
         & \makecell[l]{80.8 \\ 82.3 \gainbox{lightgreen}{$\uparrow$1.5}} \\
\cmidrule(lr){2-12}
    & JA & \makecell{$\times$ \\ \checkmark} 
         & \makecell[l]{82.4 \\ 83.6 \gainbox{lightgreen}{$\uparrow$1.2}} 
         & \makecell[l]{86.3 \\ 88.3 \gainbox{lightgreen}{$\uparrow$2.0}} 
         & \makecell[l]{69.5 \\ 70.6 \gainbox{lightgreen}{$\uparrow$1.1}} 
         & \makecell[l]{65.2 \\ 66.1 \gainbox{lightgreen}{$\uparrow$0.9}} 
         & \makecell[l]{72.6 \\ 71.5 \gainbox{lightred}{$\downarrow$1.1}} 
         & \makecell[l]{84.9 \\ 85.2 \gainbox{lightgreen}{$\uparrow$0.3}} 
         & \makecell[l]{88.0 \\ 92.0 \gainbox{lightgreen}{$\uparrow$4.0}} 
         & \makecell[l]{72.0 \\ 76.0 \gainbox{lightgreen}{$\uparrow$4.0}} 
         & \makecell[l]{77.6 \\ 79.5 \gainbox{lightgreen}{$\uparrow$1.9}} \\
\bottomrule
\end{tabular}
\caption{Impact of \textbf{GUI-XLI} on reasoning-based performance. All experiments require model to generate reasoning chains before final prediction. We compare the baseline reasoning performance ($\times$) against the enhanced performance after integrating \textbf{GUI-XLI} ($\checkmark$). \textbf{GXI} denotes GUI-XLI.}
\label{tab:reasonchain}
\end{table*}
\subsection{Main Results}
Table~\ref{tab:method_comparison} presents the evaluation results on MPR-GUI-Bench before and after applying GUI-XLI. Our analysis yields three key conclusions:

\paragraph{(1) Effective Cross-Lingual Capability Transfer}
GUI-XLI significantly enhances fine-grained P\&R capabilities in non-English settings, effectively aligning them with English-level proficiency. For high-resource languages like ZH, Intern2.5VL-8B and Qwen-2.5-VL-7B-Instruct achieve absolute $\text{FPR-ACC}$ gains of 10.9\% and 6.0\%, respectively. Crucially, this improvement is consistent across lower-resource languages (e.g., TH, JA), bridging the average performance gap by 5.4\% on average.

\paragraph{(2) Data Independence and Model Generalization}
We construct GUI-XL-Memory using data entirely disjoint from MPR-GUI-Bench, ensuring that performance gains stem from transferable P\&R patterns. In addition, consistent improvements across heterogeneous open-source LVLMs demonstrate our GUI-XLI method’s robustness and broad architectural applicability.
\begin{figure}[t]
    \centering
    \includegraphics[width=1.0\columnwidth]{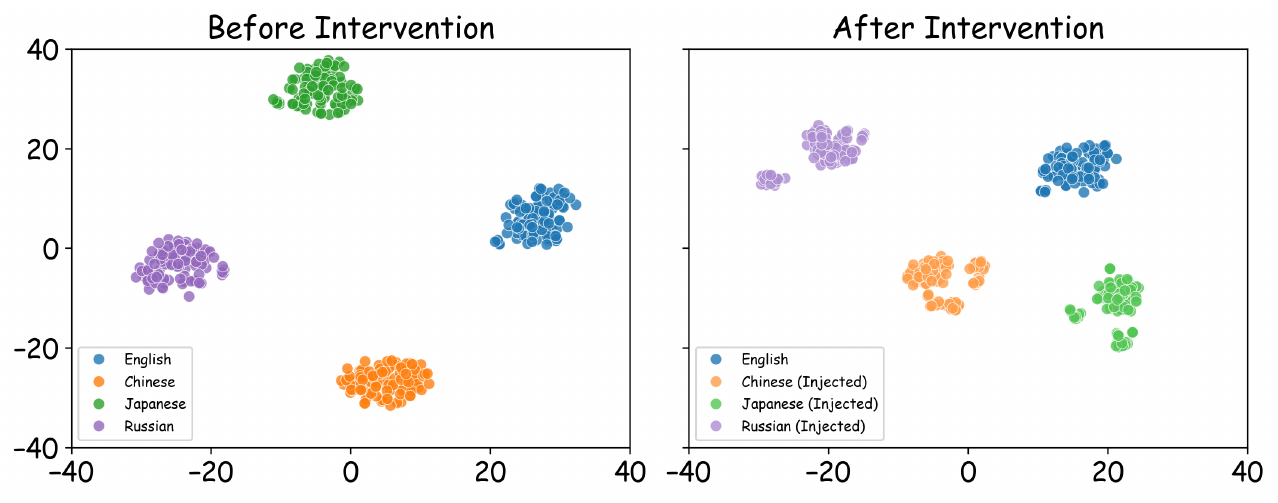}
    \caption{t-SNE Visualization of multilingual hidden state before and after applied GUI-XLI.}
    \label{fig:tsne}
\end{figure}
\paragraph{(3) Differential Gains across Task Dimensions}
Gains are most pronounced in action understanding and prediction tasks (AU, AP), with Intern2.5VL-8B and Qwen-2.5-VL-7B-Instruct achieving average $\text{FPR-ACC}$ increases of 3.7\% and 2.4\%, respectively. Conversely, simpler dimensions (e.g., WI) exhibit marginal gains due to performance saturation. Improvements in spatial and end-to-end reasoning (REL, SI) are moderate; we attribute this to their weaker dependence on linguistic context—which limits the impact of cross-lingual alignment—and their inherent complexity, which remains a bottleneck beyond language barriers.

This result suggests that our approach is particularly effective in enhancing performance in tasks that require complex reasoning and prediction, but less so for tasks that are already well-performing or involve higher-order cognitive processing.

\section{Analysis}
\label{analysis}
\subsection{Cross-lingual Alignment Visualization}
\label{Cross-lingual Alignment Visualization}
As mentioned in Section~\ref{Cross-lingual Discrepancy Vector Construction}, where intermediate layers are shown to serve as English-centric reasoning hubs, cross-lingual distributional differences in intermediate layers $h^{(l)}$ reflect discrepancies in GUI P\&R capabilities across languages. 
To gain mechanistic insight into how GUI-XLI improves non-English P\&R capabilities,
we apply t-SNE to the final-token's hidden states at the intervention layer $l$ for English inputs and their non-English counterparts (ZH, RU, JA), before and after applying GUI-XLI. As shown in Figure~\ref{fig:tsne}, without GUI-XLI, representations form distinct language-specific clusters, indicating substantial cross-lingual divergence. After applying GUI-XLI, non-English representations become more concentrated and aligned with their English counterparts, providing qualitative evidence that GUI-XLI effectively bridges cross-lingual GUI P\&R gaps.

\begin{figure}[t]
    \centering
    \includegraphics[width=1.0\columnwidth]{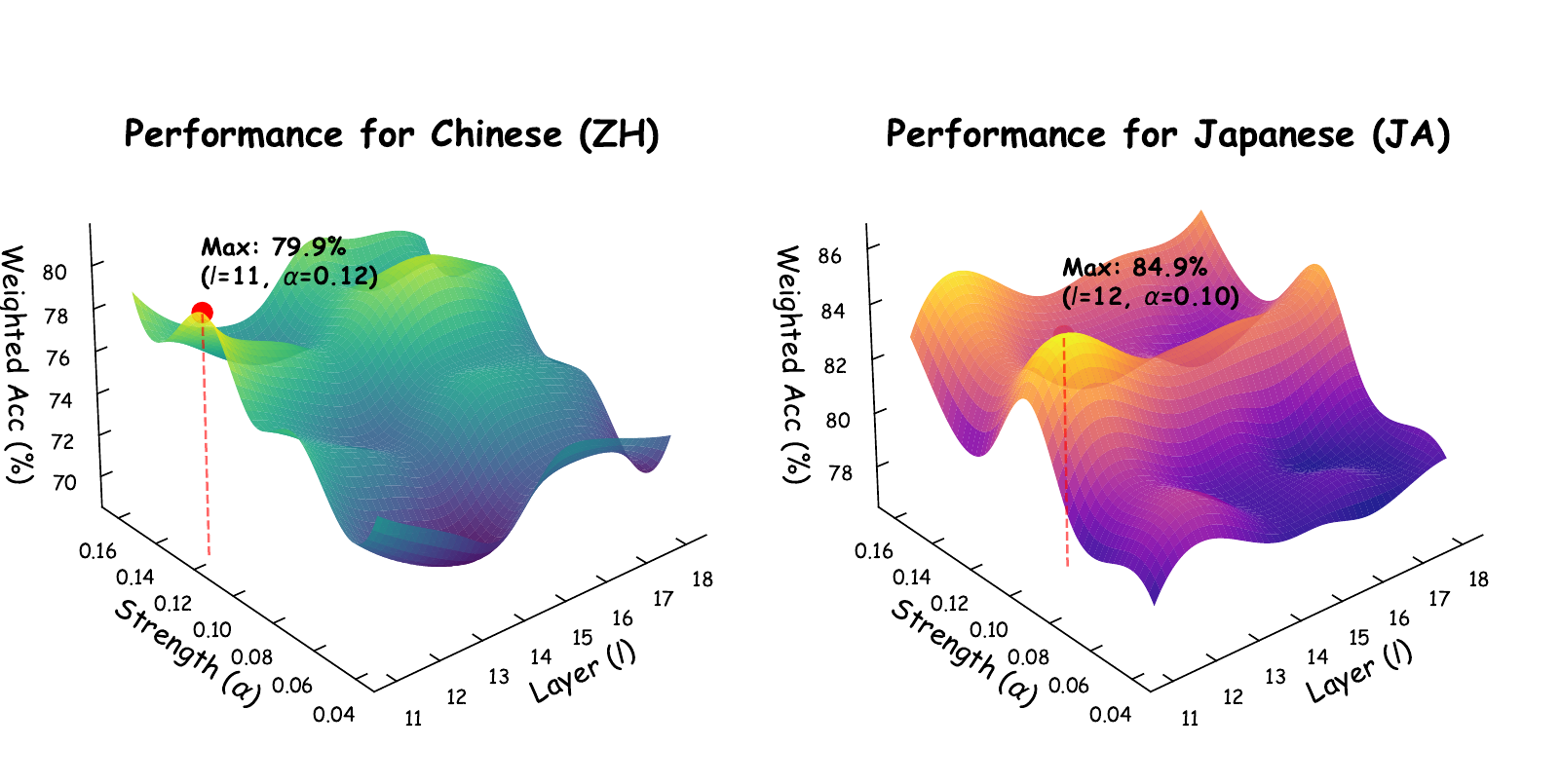}
    \caption{3D performance landscape on MPR-GUI-Bench. Variations in weighted average accuracy are shown across intervention layers $l$ and strengths $\alpha$ for Chinese and Japanese. Red dot markers denote the global optima, identifying the sweet spot for model performance in both language settings.}
\label{fig:MPR_3D_Clean_Dot_Style}
\end{figure}
\subsection{Ablation Studies}

Figure~\ref{fig:MPR_3D_Clean_Dot_Style} presents a systematic study to identify \textbf{P\&R-dominant layers} and determine the appropriate intervention strength $\alpha$. Motivated by our analysis in Section~\ref{Cross-lingual Alignment Visualization}, which shows that intermediate layers function as an English-centric reasoning hub, we believe that cross-lingual P\&R discrepancies are most salient in this layer range. We conduct a two-stage ablation. First, fixing $\alpha = 0.1$, we sweep the intervention layer $l$ across intermediate layers (11–18) to identify the optimal layer $l^{*}$. Second, conditioned on $l^{*}$, we vary $\alpha$ to assess sensitivity to intervention strength. For each model and language setting, we select the configuration yielding the highest weighted accuracy. 

\subsection{Impact on Reasoning-based Performance}
Table~\ref{tab:reasonchain} shows that beyond improving non-English P\&R performance, GUI-XLI also yields substantial gains when required to generate reasoning chains before answering. Notably, GUI-XLI consistently improves performance even under explicit reasoning supervision, indicating that the gains arise from a genuine enhancement of underlying P\&R capability rather than prompting effects. From a mechanistic perspective, reasoning chains externalize internal P\&R trajectories; the improved performance therefore suggests that GUI-XLI effectively reshapes non-English P\&R patterns toward more coherent and task-relevant reasoning.
\section{Conclusion}
In this paper, we introduce MPR-GUI-Bench, the first benchmark designed to evaluate the fundamental fine-grained perception and reasoning (P\&R) capabilities in GUI agents across strictly aligned cross-lingual environments spanning six languages. Our evaluation of current LVLMs reveals a consistent performance bottleneck in non-English settings, particularly in reasoning-intensive tasks. To leverage the superior P\&R capabilities of English, we identify critical layers that are most sensitive to linguistic discrepancies. Based on these insights, we propose GUI-XLI, an inference-time intervention method that aligns non-English representations with superior English counterparts at these critical layers. Experimental results demonstrate that GUI-XLI significantly bridges the cross-lingual gap with an absolute average performance gain of 6.5\% in non-English settings.

\section*{Limitations}
Due to limited resources, \textbf{MPR-GUI-Bench} only include mobile device models. 
Additionally, while MPR-GUI-Bench effectively evaluates the fine-grained P\&R capabilities of closed-source LVLMs, GUI-XLI cannot be extended to these models because their internal hidden representations are inaccessible for the required interventions. 
\section*{Ethical Considerations}
Our data collection follows ethical guidelines. For human annotation, all participants were informed of the research goals, participated voluntarily, and received fair compensation. All GUI screenshots and trajectories were manually de-identified to remove personally identifiable information. For data sourcing, we strictly adhered to the platforms' Terms of Service and Robots.txt. All interactions were limited to public interfaces at a controlled frequency for non-commercial academic use.
\section*{Acknowledgements}
Xiaocheng Feng is the co-corresponding author of this work.
We thank the anonymous reviewers for their insightful comments.
This work was supported by the National Natural Science Foundation of China (NSFC) (grant 62522603, 62276078), the Key R\&D Program of Heilongjiang via grant 2022ZX01A32, the Fundamental Research Funds for the Central Universities ( XNJKKGYDJ2024013 ) .
\bibliography{custom}
\clearpage
\appendix

\section{Additional Details of MPR-GUI-Bench}

\subsection{Annotator Background}
\label{appendix:Annotator Background}
The annotation process was conducted by one of the authors and five trained volunteers, all of whom hold at least a bachelor's degree and possess expertise in GUI applications. To ensure high-quality multilingual dataset construction, we implemented a \textbf{Lead-Supporting Annotator} framework. For each language, we ensured that at least two \textbf{Lead Annotators} were assigned based on their certified proficiency (e.g., TOEFL, HSK, or JLPT), with their specific roles detailed in Table~\ref{tab:annotator_proficiency}. 
\begin{table*}[t]
\centering
\begin{tabular}{ccccccc}
\toprule
Annotator & EN & ZH & FR & RU & JA & TH \\
\midrule
1 & Lead & Lead & Lead & -- & -- & -- \\
2 & Lead & Lead & -- & Lead & -- & -- \\
3 & Lead & -- & Lead & -- & Lead & -- \\
4 & Lead & -- & -- & -- & -- & Lead \\
5 & Lead & Lead & -- & -- & Lead & -- \\
6 & Lead & -- & Lead & -- & -- & Lead \\
\midrule
\textbf{Role} & Expert & Expert/Tool & Expert/Tool & Expert/Tool & Expert/Tool & Expert/Tool \\
\bottomrule
\end{tabular}
\caption{Annotator roles and language distribution. ``Lead'' indicates annotators with certified proficiency who provided direct linguistic judgments. Supporting annotators (denoted by tool-assisted roles) utilized back-translation for verification.}
\label{tab:annotator_proficiency}
\end{table*}

To prioritize annotator well-being and maintain high data quality, the \textbf{workload was strictly controlled} at 1--2 hours per day over a period of two weeks. Annotators were compensated at an average rate of 15 USD per hour, following a comprehensive training phase that included task-specific tutorials and alignment on quality criteria. 
\paragraph{Privacy and Content Safety.}
During data collection, annotators were explicitly instructed to avoid collecting any GUI content that contains personal or sensitive information. In particular, screenshots and annotations that name or uniquely identify individual people (e.g., real names, user IDs, email addresses, phone numbers, or account-related information) were excluded. After collection, all samples were manually reviewed, and any content containing personally identifiable information or potentially offensive material was filtered out. As a result, the final dataset does not include personal identifiers or offensive content.
\subsection{Data Collection Guidelines}
\label{appendix:Data Collection Guidelines}
As shown in Table~\ref{tab:DataCollectionGuidelines}, we provide guidelines for annotators on data collection and verification to ensure data quality and consistency across annotators.
\begin{table*}[htbp]
    \centering
    \begin{tabularx}{\textwidth}{X}
        \toprule
        \textbf{Data Collecting Guidelines} \\
        \midrule
        Annotators are required to collect screenshots in the following languages: Chinese (ZH), English (EN), French (FR), Russian (RU), Thai (TH), and Japanese (JA). \\
        \begin{enumerate}[label=\alph*.]
            \item First, check whether each app/website supports the above languages.
        
            \item Select an app and begin capturing screenshots. For each app/website, capture as many different screens as possible, each corresponding to one of the six language environments listed above. Try to ensure that the screens represent different scenarios.
        
            \item Next, to maintain consistency, check the initial screenshots based on the following three guidelines:
                \begin{enumerate}[label=\roman*.]
                    \item It is recommended to select as many screenshots as possible, as many might be discarded after checking. Ensure that the final dataset has at least 10 different scenes for each app.
                    
                    \item Consistency must be maintained, meaning that apart from the language, the visual style, background coherence, and text formatting should remain consistent. For example, in a weather app, the temperature unit should be the same across different languages. If the app includes recommended content (e.g., search recommendations in a browser), ensure that the recommendations remain consistent when switching languages. Additionally, if searching within a browser, the search terms should be translated according to the language (e.g., searching "apple" in English should correspond to "pingguo" in Chinese). Make sure the input method is set to the correct language as well.
                    
                    \item An example of a valid scenario is as follows: ...the 6 screenshots from the same scenario show only differences in language, while the layout remains almost identical. Such data should be retained. An example of invalid data that should be discarded: the screenshots show significant issues that hinder interface comprehension, such as inconsistent text content across languages, mixed languages, and layout issues that obstruct understanding.
                \end{enumerate}
        
            \item After checking, use GPT-4o to automatically generate questions. The recommended prompt template can be found at the end of the task instructions.
        
            \item After generating the questions, manually review and check them based on the following aspects, then either ask GPT-4o to regenerate the questions or design them manually:
                \begin{enumerate}[label=\roman*.]
                    \item For question design, check out Q\&A pairs that are factually incorrect, have mismatched objects/screenshots, or have questions that are too easy or too difficult.
                    
                    \item For option design, check out Q\&A pairs where the correct option is inaccurate, incorrect options lack sufficient distractor quality, or the options are misleading.
                \end{enumerate}
                
            \item Next, annotators need to input the semantically parallel non-English screenshots along with the translation prompts into GPT-4o, allowing the model to translate all the candidate VQA pairs from the English list into the target languages.
        
            \item Annotators must check each translated VQA pair in the target language to ensure cross-lingual consistency. If any discrepancies are found, the translation should be redone or the example discarded. This process will result in the creation of a complete dataset.
        \end{enumerate} \\
        \bottomrule
    \end{tabularx}
    \caption{Data Collecting Guidelines}
    \label{tab:DataCollectionGuidelines}
\end{table*}
\subsection{Prompts For Candidate VQA Lists Construction}
\label{apdx:Prompts For Candidate VQA Lists Construction}
In this section, we list all prompts used during the
process of constructing \textbf{MPR-GUI-Bench},
which include VQA generation for eight dimensions (Table~\ref{tab:prompt_action_understanding} - Table~\ref{tab:prompt-rISI}). Note that
for RI and SI demensions, we ask annotators to provide the goal for the screenshot sequences, so the corresponding prompt requires GPT-4o to only generate distractors.
\begin{table*}[htbp]
    \centering    
    \begin{tabularx}{\textwidth}{X}
        \toprule
        
        \textbf{Prompt for AU Dimension} \\
        \midrule
        
        You are an AI visual assistant. You are given a single screenshot captured from a mobile UI during user interaction. 
        Your task is to design ONE multiple-choice question that evaluates the model’s \textbf{Action Understanding Ability}, defined as the ability to:
        \newline\newline
        Predict the immediate outcome and effects of performing a specific action given the current interface state. Focus on:
        \newline
        1. \textbf{Interface state changes} (e.g., navigating to a new page, opening or closing a popup, expanding or collapsing content areas, toggling an icon’s opacity or color)
        \newline
        2. \textbf{Data state changes} (e.g., saving data, deleting an item)
        \newline
        3. \textbf{System feedback} (e.g., displaying a success message, an error warning, or a loading indicator)
        \newline
        4. \textbf{Impact on subsequent flow} (e.g., unlocking the next step, resetting to the initial state, reaching a terminal page, expiring a critical condition)
        \newline\newline
        \textbf{Notice}: Ensure that the questions you design for these tasks are answerable and the answers can be deduced from the GUI content.
        You must make each question as \textbf{difficult and nuanced} as possible, requiring careful visual perception and contextual reasoning. Avoid obvious or overly simple options. Include plausible distractors for each question to increase the difficulty.
        \newline\newline
        For each given screenshot, create one multiple-choice question that tests one of the abilities mentioned above. Each question should have four answer options: one correct answer and three that are incorrect but closely relevant. Distractors should be designed to be tempting yet contain subtle mistakes drawn from the interface that are difficult to detect.
        \newline\newline
        Your reply must be structured like this, with \textbf{no extra explanation}:
        \newline\newline
        question: \{your question\}
        \newline
        options:
        \newline
        A. \{Option A\}
        \newline
        B. \{Option B\}
        \newline
        C. \{Option C\}
        \newline
        D. \{Option D\}
        \newline
        answer: \{Correct option letter\}
        \newline
        type: Action Understanding
        \newline\newline
        Please keep each answer as \textbf{concise and difficult} as possible, and only structured in this exact format. Only include questions that you can answer confidently based on the image content.
        \\ 
        
        \bottomrule
    \end{tabularx}
     \caption{Prompt for AU Dimension}
     \label{tab:prompt_action_understanding}
\end{table*}

\begin{table*}[htbp]
    \centering

    \begin{tabularx}{\textwidth}{X}
        \toprule
        
        \textbf{Prompt for AP Dimension} \\
        \midrule
        
        You are an AI visual assistant. You are given a single screenshot captured from a mobile UI during user interaction.
        \newline\newline
        Below, you will be provided with a hypothetical user task goal. Your task is to design ONE multiple-choice question that evaluates the model’s \textbf{Action Prediction Ability}, defined as the ability to:
        \newline\newline
        1. \textbf{Action Type}: Select the correct interaction type from the set \{tap, long press, swipe, type text, press home/back/recent\}.
        \newline
        2. \textbf{Action Target}: Identify the precise UI element to interact with.
        \newline
        3. \textbf{Input Content}: If text input is required, specify the exact text.
        \newline
        4. \textbf{Action Sequence}: For multi-step tasks, determine the correct order of operations.
        \newline\newline
        Your question must:
        \newline
        \hspace*{1em} - \textbf{Embed} a clear user task goal (e.g., “The user wants to add a new contact with name X and phone Y”).
        \newline
        \hspace*{1em} - Ask: “To achieve this goal, which of the following description is true?“
        \newline
        \hspace*{1em} - Provide \textbf{four} answer options (A–D), at least one option should describe a full sequence of actions, it could be the correct one or a distractor.
        \newline
        \hspace*{1em} - \textbf{One} correct sequence.
        \newline
        \hspace*{1em} - \textbf{Three} distractors that each violate at least one of:
        \newline
        \hspace*{2em} - Wrong action type on a step.
        \newline
        \hspace*{2em} - Missing a critical step.
        \newline
        \hspace*{2em} - Steps in the incorrect order.
        \newline
        \hspace*{2em} - Wrong target element.
        \newline
        \hspace*{1em} - Make options concise but \textbf{nuanced}—avoid obvious mistakes.
        \newline\newline
        \textbf{Structure your reply with NO extra text}:
        \newline\newline
        question: \{your question embedding the user goal\}
        \newline
        options:
        \newline
        A. \{step1 $\rightarrow$ step2 $\rightarrow$ \dots\}
        \newline
        B. \{\dots\}
        \newline
        C. \{\dots\}
        \newline
        D. \{\dots\}
        \newline
        answer: \{Correct option letter\}
        \newline
        type: Action Prediction
        \\ 
        
        \bottomrule
    \end{tabularx}
        \caption{Prompt for AP Dimension}
    \label{tab:prompt_action_prediction}
\end{table*}

\begin{table*}[htbp]
    \centering
    \label{tab:ael_PROMPT}

    \begin{tabularx}{\textwidth}{X} 
        \toprule
        
        \textbf{Prompt for AEL Dimension} \\
        \midrule
        
        You are an AI visual assistant. You are given a single screenshot captured from a mobile UI during user interaction.
        Your task is to design one multiple-choice question that evaluates the \textbf{Absolute Element Location Ability}. Specifically, you should strictly follow these guidelines:
        \newline\newline
        1. \textbf{Question Description}:
        \newline
        - Clearly specify the element to be located (e.g., "Please determine the position of the blue button on the screen").
        \newline
        - Ask the model to analyze the general area of this element within the global coordinate system.
        \newline\newline
        2. \textbf{Reference Layout Structure}:
        \newline
        - Prompt the model to consider the overall interface structure (e.g., top navigation bar, central content area, bottom action bar) when making its determination.
        \newline
        - Guide the model to identify which section the element belongs to, such as status bar / toolbar / main content area / floating button area.
        \newline\newline
        3. \textbf{Absolute Position Description}:
        \newline
        - Require the model to use standardized regions:
        \newline
        \hspace*{1em} - Quadrant-based description: upper-left / lower-left / upper-right / lower-right;
        \newline
        \hspace*{1em} - Alternatively, a three-part division: top / middle / bottom.
        \newline
        - The question stem or options must explicitly use the above-mentioned descriptive terms to clearly define the location.
        \newline\newline
        \textbf{Notice}: Ensure that the questions you design for these tasks are answerable and the answers can be deduced from the GUI content.
        You must make each question as \textbf{difficult and nuanced} as possible, requiring careful visual and contextual reasoning. Avoid obvious or overly simple options. Minimize the repetition of the questioned objects as much as possible. Include plausible distractors for each question to increase the difficulty.
        \newline\newline
        For each given screenshot, create one multiple-choice question that tests one of the abilities mentioned above. Each question should have four answer options: one correct answer and three that are incorrect or irrelevant.
        \newline\newline
        Your reply must be structured like this, with \textbf{no extra explanation}:
        \newline\newline
        question: \{your question\}
        \newline
        options:
        \newline
        A. \{Option A\}
        \newline
        B. \{Option B\}
        \newline
        C. \{Option C\}
        \newline
        D. \{Option D\}
        \newline
        answer: \{Correct option letter\}
        \newline
        type: Absolute Element Location
        \newline\newline
        Please keep each answer as \textbf{concise and focused} as possible, and only include the questions in this exact format. Only include questions that have definite answers.
        \\ 
        
        \bottomrule
    \end{tabularx}
        \caption{Prompt for AEL Dimension}
\end{table*}

\begin{table*}[htbp]
    \centering
    \label{tab:prompt_relative_loc}

    \begin{tabularx}{\textwidth}{X}
        \toprule
        
        \textbf{Prompt for REL Dimension} \\
        \midrule
        
        You are an AI visual assistant. You are given a single screenshot captured from a mobile UI during user interaction.
        Your task is to design one multiple-choice question that evaluates the \textbf{Relative Element Location Ability}. This refers to evaluating the model's ability to reason about spatial relationships in the interface. Specifically, the model’s ability to:
        \newline\newline
        \hspace*{1em} - Determine the relative location of elements on the interface.
        \newline\newline
        \textbf{Notice}: Ensure that the questions you design for these tasks are answerable and the answers can be deduced from the GUI content.
        You must make each question as \textbf{difficult and nuanced} as possible, requiring careful visual and contextual reasoning. Avoid obvious or overly simple options. Minimize the repetition of the questioned objects as much as possible. Include plausible distractors for each question to increase the difficulty.
        \newline\newline
        For each given screenshot, create one multiple-choice question that tests one of the abilities mentioned above. Each question should have four answer options: one correct answer and three that are incorrect or irrelevant.
        \newline\newline
        Your reply must be structured like this, with \textbf{no extra explanation}:
        \newline\newline
        question: \{your question\}
        \newline
        options:
        \newline
        A. \{Option A\}
        \newline
        B. \{Option B\}
        \newline
        C. \{Option C\}
        \newline
        D. \{Option D\}
        \newline
        answer: \{Correct option letter\}
        \newline
        type: Relative Element Location
        \newline\newline
        Please keep each answer as \textbf{concise and focused} as possible, and only include the questions in this exact format. Only include questions that have definite answers.
        \\ 
        
        \bottomrule
    \end{tabularx}
        \caption{Prompt for REL Dimension}
\end{table*}
\begin{table*}[htbp]
    \centering
    \label{tab:prompt_widget_perception}

    \begin{tabularx}{\textwidth}{X}
        \toprule
        
        \textbf{Prompt for WF Dimension} \\
        \midrule
        
        You are an AI visual forensics analyst specializing in mobile UI screenshots. Design ONE expert-level multiple-choice question that rigorously tests \textbf{Widget Function Perception Ability} with these constraints:
        \newline\newline
        \textbf{Strict visual evidence requirements:}
        \newline
        \hspace*{1em} - All answers MUST be provable from explicit visual evidence
        \newline
        \hspace*{1em} - Absolutely NO speculation beyond what's visible
        \newline
        \hspace*{1em} - Correct answers require synthesizing $\geq$3 distinct visual cues
        \newline
        \hspace*{1em} - Reject any interpretation not confirmed by:
        \newline
        \hspace*{2em} 1. Standard platform conventions
        \newline
        \hspace*{2em} 2. Explicit visual affordances (shadows, highlights, depth cues)
        \newline
        \hspace*{2em} 3. State indicators (color coding, iconography, text labels)
        \newline
        \hspace*{2em} 4. Spatial relationships to adjacent elements
        \newline\newline
        \textbf{Core ability focus (evidence-based):}
        \newline
        \hspace*{1em} - MUST synthesize $\geq$3 distinct visual cues:
        \newline
        \hspace*{2em} 1. Primary text labels (e.g., "Weather", "Reminders")
        \newline
        \hspace*{2em} 2. Icon semantics (standard meanings only)
        \newline
        \hspace*{2em} 3. Data representations (charts, progress bars)
        \newline
        \hspace*{2em} 4. Contextual positioning (status bar vs. home screen)
        \newline
        \hspace*{1em} - BANNED:
        \newline
        \hspace*{2em} 1. Speculation beyond visible elements
        \newline
        \hspace*{2em} 2. Prior knowledge of specific apps
        \newline\newline
        \textbf{Question design requirements:}
        \newline
        \hspace*{1em} - Ambiguous but decodable visual patterns (e.g., semi-transparent overlay on a search icon requiring icon shape, faded color, and nearby label)
        \newline
        \hspace*{1em} - Compound state indicators (e.g., lock icon + greyed-out button requiring icon meaning and color state)
        \newline
        \hspace*{1em} - Conflicting affordances requiring prioritization (e.g., send arrow and trash icon in the same area)
        \newline
        \hspace*{1em} - are platform-specific edge cases (e.g., Android 11 share button long-press reveals hidden menu)
        \newline\newline
        \textbf{Your reply must be structured like this, with no extra explanation:}
        \newline\newline
        question: \{your question\}
        \newline
        options:
        \newline
        A. \{Option A\}
        \newline
        B. \{Option B\}
        \newline
        C. \{Option C\}
        \newline
        D. \{Option D\}
        \newline
        answer: \{Correct option letter\}
        \newline
        type: Widget Function
        \\ 
        
        \bottomrule
    \end{tabularx}
        \caption{Prompt for WF Dimension}
\end{table*}
\begin{table*}[htbp]
    \centering
    \label{tab:prompt_widget_interaction}

    \begin{tabularx}{\textwidth}{X}
        \toprule
        
        \textbf{Prompt for WI Dimension} \\
        \midrule
        
        You are an AI visual assistant. You are given a single screenshot captured from a mobile UI during user interaction.
        Your task is to design ONE multiple-choice question that evaluates the model’s \textbf{Widget Interaction Perception Ability}, defined as inferring how users can interact with visible widgets by analyzing the given mobile UI screenshot. Specifically, the ability to:
        \newline\newline
        1. \textbf{Identify Interactive Elements}
        \newline
        \hspace*{1em} Recognize actionable widgets (buttons, sliders, toggles, input fields, etc.) and distinguish them from static elements.
        \newline
        2. \textbf{Predict Interaction Methods}
        \newline
        \hspace*{1em} Determine valid operation types for each widget (tap, double-tap, long-press, swipe, pinch, etc.).
        \newline
        3. \textbf{Anticipate Interaction Outcomes}
        \newline
        \hspace*{1em} Foresee the immediate results of interactions, including:
        \newline
        \hspace*{2em} - Interface transitions (e.g., opening a settings panel)
        \newline
        \hspace*{2em} - State changes (e.g., toggle switching)
        \newline
        \hspace*{2em} - Function executions (e.g., alarm creation)
        \newline
        4. \textbf{Understand Practical Utility}
        \newline
        \hspace*{1em} Explain how the interaction solves real-world problems or enhances convenience, such as:
        \newline
        \hspace*{2em} - "Clicking '+' on clock widget enables quick alarm setting"
        \newline
        \hspace*{2em} - "Swiping down the corner slider adjusts screen brightness"
        \newline
        \hspace*{2em} - "Tapping screen time widget reveals detailed usage analytics"
        \newline\newline
        \textbf{Notice}: Ensure that the questions you design for these tasks are answerable and the answers can be deduced from the GUI content.
        You must make each question as difficult and nuanced as possible, requiring careful visual perception and contextual reasoning. Avoid obvious or overly simple options. Include plausible distractors for each question to increase the difficulty.
        For each given screenshot, create one multiple-choice question that tests one of the abilities mentioned above. Each question should have four answer options: one correct answer and three that are incorrect but closely relevant. Distractors should be designed to be tempting yet contain subtle mistakes drawn from the interface that are difficult to detect.
        The options should include at least one non-interactive distractor (static element misuse). Your reply must be structured like this, with \textbf{no extra explanation}:
        \newline\newline
        question: \{your question\}
        \newline
        options:
        \newline
        \hspace*{1em} A. \{Option A\}
        \newline
        \hspace*{1em} B. \{Option B\}
        \newline
        \hspace*{1em} C. \{Option C\}
        \newline
        \hspace*{1em} D. \{Option D\}
        \newline
        answer: \{Correct option letter\}
        \newline
        type: Widget Interaction
        \newline\newline
        Please keep each answer as concise and difficult as possible, and only structured in this exact format. Only include questions that you can answer confidently based on the image content.
        \\ 
        
        \bottomrule
    \end{tabularx}
        \caption{Prompt for WI Dimension}
\end{table*}
\begin{table*}[htbp]
    \centering

    \begin{tabularx}{\textwidth}{X}
        \toprule
        
        \textbf{Prompt for RI \& SI Dimensions} \\
        \midrule
        
        You are an AI assistant generating multiple-choice questions to evaluate understanding of mobile UI task flows.
        \newline\newline
        The following screenshots capture a short interaction sequence in a mobile app.
        \newline\newline
        The correct user goal is:
        \newline
        "\{correct\_goal\}"
        \newline\newline
        Your task is to generate \textbf{three incorrect but plausible alternative user goals} that could reasonably be mistaken for what the user is trying to do, based on the visual context.
        \newline\newline
        \textbf{Guidelines}:
        \newline\newline
        1. Each option should \textbf{look like a real user task} — it doesn't need to match the exact phrasing or grammar of the correct goal, but should feel natural and fit within the app's context (e.g., settings, messaging, shopping, file management).
        \newline
        2. Focus on \textbf{plausible misinterpretations}: the user might think the person is doing something related but different — changing a setting instead of deleting, sharing instead of saving, searching for a contact instead of calling, etc.
        \newline
        3. Vary the \textbf{action}, \textbf{target}, or \textbf{intent}: use different verbs (edit, find, enable, share, create, view, check, etc.) or objects (a message, a photo, an account, a notification, etc.) that appear or could appear in the interface.
        \newline
        4. It’s okay if the grammar is slightly informal or simplified — real users don’t always phrase tasks perfectly.
        \newline
        5. Do \textbf{not} include explanations, reasoning, or meta-comments (e.g., no “attempt to”, “mistake”, “analyze”).
        \newline
        6. Make sure the options are clearly different from the correct goal, but still \textbf{contextually grounded} in the screenshots.
        \newline\newline
        Only output the three distractors in the following format:
        \newline\newline
        A. ...
        \newline
        B. ...
        \newline
        C. ...
        \\ 
        
        \bottomrule
    \end{tabularx}
    \caption{Prompt for RI \& SI Dimensions}
    \label{tab:prompt-rISI}

\end{table*}
\subsection{Validation on GPT-4o Translation}
\label{appendix:Validation on GPT-4o Translation}
To validate the translation quality of GPT-4o, we adopt the back translation method. First, we randomly sample 500 English VQAs from our MPR-GUI-Bench. Then we leverage GPT-4o to translate these questions to other 5 languages according to \~ref{step 4}, followed by translating them back to English. Finally, we evaluate the accuracy of Qwen 2.5VL-7B-Instruct on these samples and the evaluation result is present in Table~\ref{tab:backtrnas}.
\begin{table}[h!]
\centering
\begin{tabular}{cc}
\toprule
\textbf{Translation Path} & \textbf{Accuracy (\%)} \\
\midrule
Original (EN)        & 87.2\\
ZH \(\rightarrow\) EN & 87.2 \\
JA \(\rightarrow\) EN & 86.6 \\
RU \(\rightarrow\) EN & 86.0 \\
FR \(\rightarrow\) EN & 87.0 \\
TH \(\rightarrow\) EN & 86.2 \\
\bottomrule
\end{tabular}
\caption{Back-translation Accuracy (\%) of Qwen 2.5VL-7B-Instruct on 500 VQA Samples. The first column shows accuracy on original English questions, while subsequent columns show accuracy on questions back-translated from the target language to English.}
\label{tab:backtrnas}
\end{table}
\subsection{Comprehensive Inter-Rater Reliability Analysis for All Languages}
\label{appendix:Comprehensive Inter-Rater Reliability Analysis for All Languages}

We report the inter-rater reliability analysis for all six languages included in MPR-GUI-Bench (EN, ZH, FR, RU, JA, TH). In the English setting, annotators verified the compliance of generated questions with task requirements. For the remaining languages, the evaluation focused on the linguistic faithfulness of translated text relative to the visual content in the respective screenshots. This analysis covers all 2,156 samples per language, with six annotators classifying each as ``Compliant'' or ``Non-compliant.''

The annotators who have proficiency in respective languages provided direct judgments on the faithfulness of GPT-4o translations relative to the visual content. Other annotators served as \textbf{Supporting Annotators}, utilizing auxiliary tools (e.g., DeepL and Google Translate) to translate model outputs back into English and cross-reference them with the original visual prompts for auxiliary semantic verification. 

\paragraph{Distribution of Rater Agreement.}
Table \ref{tab:all_languages_distribution} presents the distribution of rater agreement across all languages. Across the entire benchmark, a high level of consensus was observed, with the majority of samples achieving perfect (6:0) or near-perfect (5:1) agreement. Notably, the high proportion of ``6 vs. 0'' cases confirms the effectiveness of our Lead-Supporting annotator framework and the clarity of our annotation guidelines.

\begin{table}[ht]
\centering
\small
\begin{tabular}{lcccccc}
\toprule
\textbf{Agmt.} & \textbf{EN} & \textbf{ZH} & \textbf{FR} & \textbf{RU} & \textbf{JA} & \textbf{TH} \\
\midrule
6 vs. 0 & 1693 & 1650 & 1621 & 1634 & 1602 & 1588 \\
5 vs. 1 & 291  & 312  & 340  & 325  & 355  & 360  \\
4 vs. 2 & 110  & 120  & 135  & 128  & 140  & 145  \\
3 vs. 3 & 42   & 54   & 45   & 50   & 48   & 52   \\
2 vs. 4 & 16   & 15   & 12   & 14   & 10   & 9    \\
1 vs. 5 & 1    & 3    & 2    & 3    & 1    & 2    \\
0 vs. 6 & 3    & 2    & 1    & 2    & 0    & 0    \\
\midrule
\textbf{Total} & 2156 & 2156 & 2156 & 2156 & 2156 & 2156 \\
\bottomrule
\end{tabular}
\caption{Summary of Rater Agreement Distribution for all six languages. \textbf{Agmt.} denotes the Agreement level; numbers represent the count of items for each agreement configuration (e.g., ``6 vs. 0'' denotes total consensus).}
\label{tab:all_languages_distribution}
\end{table}

\paragraph{Comparative Reliability Metrics.}
As shown in Table \ref{tab:reliability_metrics_summary}, the Fleiss' Kappa coefficients across all languages fall within a relatively low range (0.15--0.22). This is a well-documented phenomenon known as the \textit{prevalence paradox} \citep{cicchetti1990high}, which occurs when the distribution of categories is highly skewed. In our case, since over 94\% of samples are classified as ``Compliant,'' the probability of agreement by chance ($\bar{P}_e$) is naturally very high. Consequently, even with a high observed agreement ($\bar{P} > 0.90$), the Kappa coefficient remains low because it only measures the marginal improvement over an already high baseline of chance.

To provide a more robust assessment that accounts for this imbalance, we also report \textbf{Gwet's AC1} coefficient, which is mathematically less sensitive to the prevalence problem. As shown in the table, our AC1 values consistently exceed 0.85 across all languages, indicating ``almost perfect'' agreement according to standard benchmarks and validating the reliability of our annotation process.
\begin{table}[ht]
\centering
\small 
\begin{tabular}{lcccc}
\toprule
\textbf{Lang.} & \textbf{$P_o$} & \textbf{$P_e$} & \textbf{Fleiss' $\kappa$} & \textbf{Gwet's AC1} \\
\midrule
EN  & 0.9120 & 0.8942 & 0.1682 & 0.8950 \\
ZH  & 0.9055 & 0.8850 & 0.1783 & 0.8892 \\
FR   & 0.8994 & 0.8790 & 0.1686 & 0.8810 \\
RU  & 0.9021 & 0.8812 & 0.1752 & 0.8845 \\
JA & 0.8950 & 0.8710 & 0.1860 & 0.8780 \\
TH    & 0.8920 & 0.8680 & 0.1818 & 0.8755 \\
\bottomrule
\end{tabular}
\caption{Summary of statistical reliability metrics for all languages. \textbf{Lang.} is Language; \textbf{$P_o$} is Observed Agreement; \textbf{$P_e$} is Expected (Chance) Agreement. The high AC1 values across all settings validate the robustness of the MPR-GUI-Bench data collection process despite the prevalence paradox impacting the $\kappa$ values.}
\label{tab:reliability_metrics_summary}
\end{table}
\paragraph{Adjudication and Quality Control.}
For each language, all samples that failed to reach a 6:0 consensus were subjected to a final \textbf{Adjudication Phase}. The respective Lead Annotators (native or proficient speakers) resolved all disagreements to ensure the final ground-truth labels were accurate. 

This rigorous multi-stage verification process guarantees that the linguistic and visual alignment of MPR-GUI-Bench remains reliable despite the inherent difficulty of multilingual GUI evaluation.

\subsection{Quality Assurance and Refinement to Eliminate Model-specific Semantic Style}
\label{appendix:Quality Assurance and Refinement to Eliminate Model-specific Semantic Style}
As mentioned in Appendix~\ref{apdx:Prompts For Candidate VQA Lists Construction}, to safeguard the professional rigor and stylistic neutrality of MPR-GUI-Bench, we developed dimension-specific constrained prompts. This framework decomposes complex GUI interactions into fundamental sub-abilities—such as requiring multi-cue synthesis for Widget Function perception and strict sequence validation for Action Prediction. By incorporating Negative Linguistic Constraints to prohibit meta-comments and conversational fillers, we minimize the stylistic artifacts of the generator, ensuring the benchmark targets genuine visual reasoning over linguistic shortcuts.

Furthermore, we implemented rigorous annotation guidelines to refine all GPT-4o-generated samples as shown in Table~\ref{tab:Quality Assurance and Refinement Guidelines}. 

To verify that MPR-GUI-Bench is generator-agnostic, we conducted a cross-model evaluation using task sets independently generated by GPT-4o and Gemini-2.5-Pro. The results, summarized in Table~\ref{tab:cross-verification on Gemini}, provide strong evidence that our evaluation pipeline is resilient to model-specific bias.
\paragraph{Absence of Generator Dominance} Empirical data show that the generator backbone does not inherently dominate its own test set. On the GPT-4o-generated subset, Qwen-2.5-VL-7B-Instruct achieves the highest scores in AEL (78\%) and WF (86\%), surpassing the generator (GPT-4o). Conversely, on the Gemini-2.5-Pro-generated subset, GPT-4o maintains a performance lead in 5 out of 8 tasks (e.g., AU, WI, WF, SI), despite the tasks being curated by Gemini. This lack of "home-field advantage" confirms that the diagnostic data represents objective P\&R challenges rather than stylistic artifacts.
\paragraph{High Ranking Consistency} Despite variations in absolute scores—likely due to differing task difficulty distributions—the relative performance hierarchy remains remarkably stable. We observe a 100\% Top-1 ranking consistency in key perception and reasoning dimensions, including AEL, REL, AU, and WI. For instance, GPT-4o consistently outranks Gemini-2.5-Pro in WI across both generation sources (94\% vs. 79\% and 92\% vs. 89\%). The addition of RI and SI tasks further reinforces this stability, with Gemini and GPT-4o maintaining shared dominance in RI (96\% and 92\% respectively) regardless of the data source.
\paragraph{Capability-Driven vs. Style-Driven Results} The robust performance of Qwen-2.5-VL-7B-Instruct, serving as a non-generator "third-party" model, provides additional validation. Qwen consistently achieves top-tier results in AEL (ranking 1st in both subsets with 78\% and 81\%), proving that the benchmark measures standardized GUI interaction skills that transcend specific LLM prompting styles.
\paragraph{Quantitative Verification} Quantitatively, the Metric-wise Ranking Concordance remains high (averaging $\tau \approx 0.70$ across tasks, with perfect $\tau = 1.0$ in AEL). Given the narrow performance margins between these state-of-the-art models (averaging $<2\%$), this degree of concordance is statistically significant. It indicates that the performance hierarchy is driven by the intrinsic P\&R capabilities of the evaluated models rather than stylistic alignment with the generator, effectively neutralizing potential self-preference bias.
\begin{table*}[htbp]
    \centering
    \begin{tabularx}{\textwidth}{X}
        \toprule
        \textbf{Quality Assurance and Refinement Guidelines} \\
        \midrule

        \textbf{1. Objective.}
        
        Ensure that each VQA sample is factually accurate, strictly grounded in visual evidence, and presents a \emph{human-solvable yet non-trivial} reasoning challenge without inheriting model-specific semantic styles. 
        If two independent annotators disagree on the answer after inspection, the sample must be flagged and revised or removed.
        \emph{Non-trivial} indicates that the question cannot be answered via a single salient cue and requires integration of at least two visual or semantic cues (e.g., icon + text, color + position). \\[4pt]

        \textbf{2. Question Evaluation Criteria.}
        
        (i) \emph{Visual Existence \& Strict Grounding}: all referenced UI elements must be clearly identifiable in the screenshot;
        
        (ii) \emph{Factual Integrity}: zero incorrect assumptions about GUI state (e.g., disabled vs.\ enabled);
        
        (iii) \emph{Moderate Complexity}: avoid trivial perception-only questions; require contextual perception and reasoning;
        
        (iv) \emph{Human Answerability \& Legibility}: discard samples with blurred, truncated elements or those requiring insider app knowledge. \\[4pt]

        \textbf{3. Answer and Distractor Standards.}
        
        The correct answer must be the \emph{only} logically sound option. Distractors must be grounded in the same interface and reflect plausible misinterpretations (e.g., similar icons in different regions), while remaining mutually exclusive. 
        Distractors must not be partially or conditionally correct and should introduce subtle logical or spatial errors that penalize superficial pattern matching. \\[4pt]

        \textbf{4. Stylistic and Terminology Refinement.}
        
        Remove AI-typical reasoning traces and conversational fillers to prevent exploitation of linguistic priors. 
        Enforce a concise, imperative tone consistent with real GUI interactions. 
        Standardize platform-specific terminology (e.g., ``Tap'' for mobile) and ensure localization fidelity by matching the exact UI lexicon shown on the screen for non-English samples. \\[4pt]

        \textbf{5. Failure Mode Filtering (Mandatory Veto).}
        
        Discard or revise samples exhibiting: 
        
        (i) spatial inconsistency (incorrect relative positioning); 
        
        (ii) interactional logic violations in action prediction tasks; 
        
        (iii) shortcut cues such as systematic length, tone, or punctuation bias in the correct option; 
        
        (iv) evidence insufficiency, where answers rely on brand- or app-specific prior knowledge rather than explicit visual evidence. \\

        \bottomrule
    \end{tabularx}
    \caption{Quality assurance and refinement guidelines used during data collection and annotation.}
    \label{tab:Quality Assurance and Refinement Guidelines}
\end{table*}
\begin{table*}[ht]
\centering
\small
\begin{tabular}{lcccccccc}
\toprule
\textbf{Evaluated Model} & \textbf{AEL} & \textbf{REL} & \textbf{AU} & \textbf{AP} & \textbf{WI} & \textbf{WF} & \textbf{RI} & \textbf{SI} \\
\midrule
\rowcolor[gray]{0.95} \multicolumn{9}{c}{\textit{Generated by GPT-4o}} \\
Gemini-2.5-Pro           & 58          & \textbf{92} & 81          & \textbf{91} & 79          & 63          & \textbf{96} & \textbf{80} \\
GPT-4o                   & 72          & 83          & \textbf{92} & \textbf{91} & \textbf{94} & 84          & \textbf{96} & 76          \\
Qwen-2.5-VL-7B-Instruct  & \textbf{78} & 90          & 84          & \textbf{91} & 91          & \textbf{86} & \textbf{96} & 72          \\
\midrule
\rowcolor[gray]{0.95} \multicolumn{9}{c}{\textit{Generated by Gemini-2.5-pro}} \\
Gemini-2.5-Pro           & 72          & \textbf{67} & 75          & \textbf{90} & 89          & 79          & \textbf{92} & \textbf{82} \\
GPT-4o                   & 77          & 60          & \textbf{80} & 88          & \textbf{92} & \textbf{86} & \textbf{92} & \textbf{82} \\
Qwen-2.5-VL-7B-Instruct  & \textbf{81} & 55          & 70          & 89          & 91          & 83          & 88          & 72          \\
\bottomrule
\end{tabular}

\caption{Cross-verification of model bias using QAs generated by different backbones.}
\label{tab:cross-verification on Gemini}

\end{table*}
\subsection{Details about $\text{FPR-ACC}$}
\label{appendix:Details about FPR-ACC}
We use the $\text{FPR-ACC}$ parameter as the comprehensive score for the fine-grained P\&R capabilities of the model on our MPR-GUI-Bench. Specifically, we categorize the eight task dimensions into three difficulty levels. The six static dimensions (Table~\ref{tab:performance_result}, d1--d6) involve only single-image perception and are assigned a base weight of $w_i = 1$. The RI dimension (d7), which benefits from temporal context across multiple screenshots, is assigned a medium weight of $w_7 = 1.5$. The SI dimension (d8), which requires inferring user intentions from minimal visual evidence and sparse information and represents the highest reasoning challenge, is assigned the largest weight of $w_8 = 2$.

\subsection{Case Study}
\label{appendix:case study}
 To gain deeper insights into the limitations of current LVLMs, we conduct a qualitative analysis of typical failure modes across the various dimensions of \textbf{MPR-GUI-Bench}. Figures~\ref{fig:ael_case} through \ref{fig:si_case} illustrate representative incorrect responses, highlighting the specific challenges they encounter in GUI perception and reasoning.

\begin{figure*}[t]
    \centering
    \includegraphics[width=2.0\columnwidth]{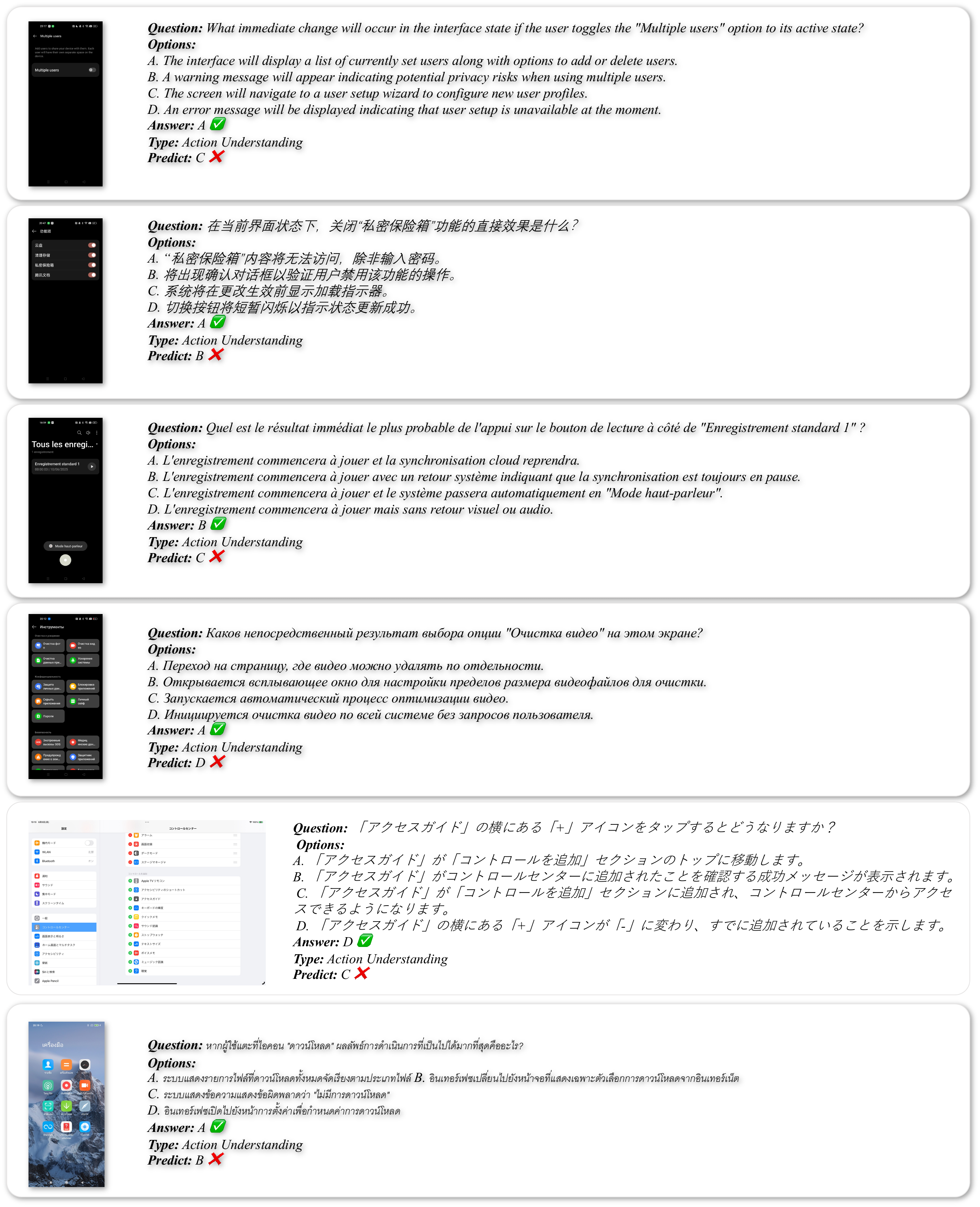}
    \caption{Examples of incorrect responses by LVLMs in AU dimension across 6 language settings.}
    \label{fig:au_case}
\end{figure*}
\begin{figure*}[t]
    \centering
    \includegraphics[width=2.0\columnwidth]{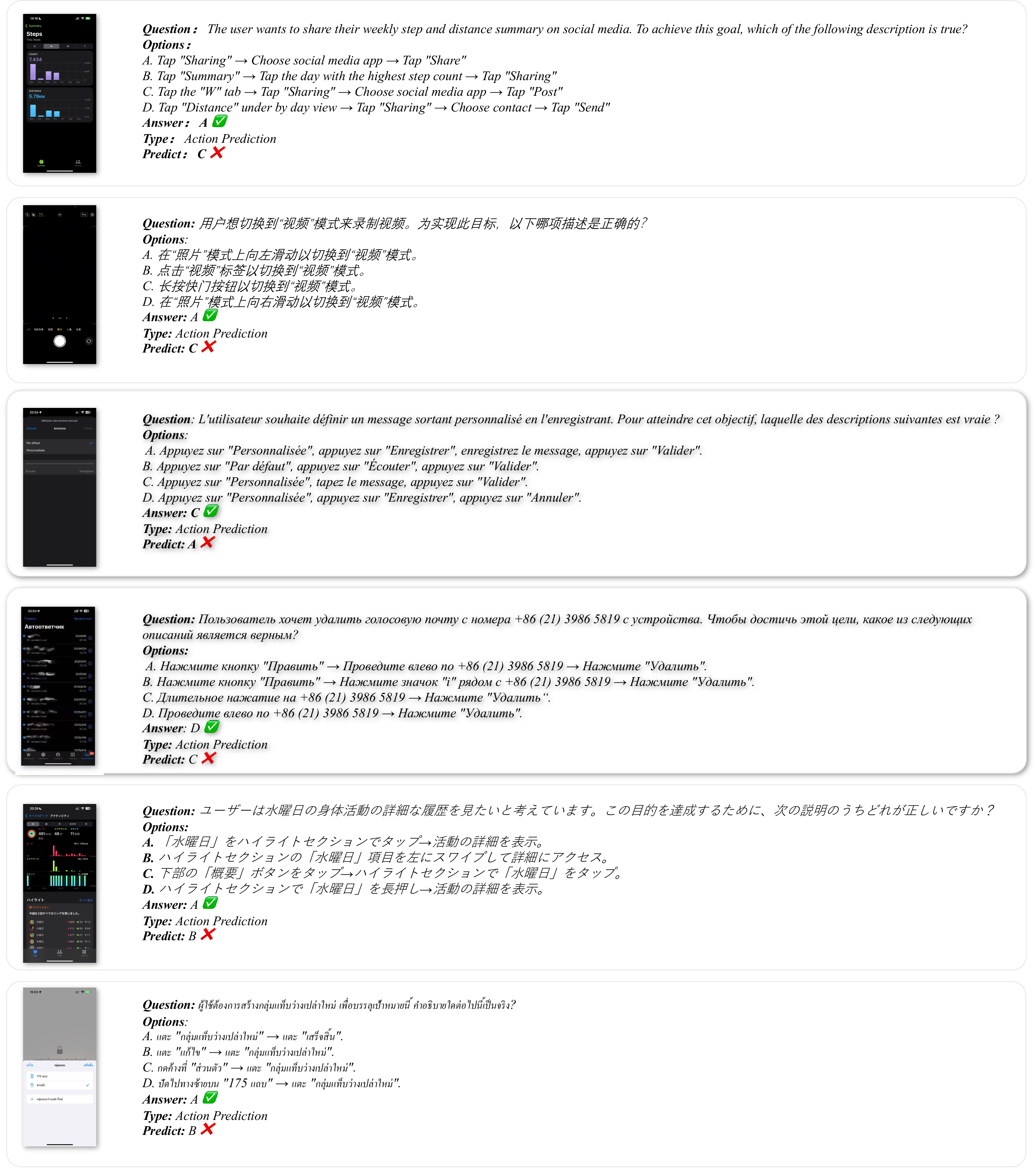}
    \caption{Examples of incorrect responses by LVLMs in AP dimension across 6 language settings.}
    \label{fig:ap_case}
\end{figure*}
\begin{figure*}[t]
    \centering
    \includegraphics[width=2.0\columnwidth]{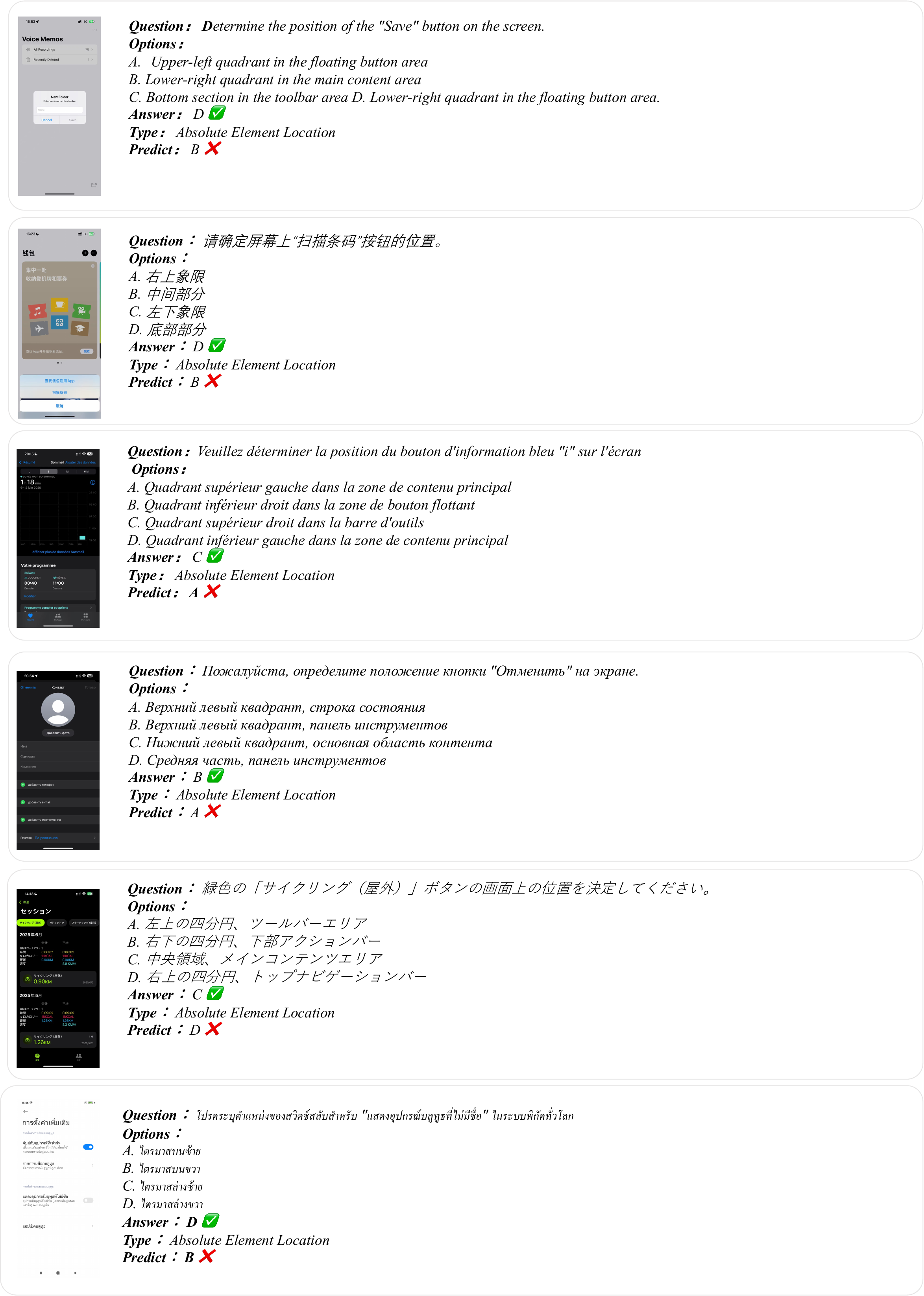}
    \caption{Examples of incorrect responses by LVLMs in AEL dimension across 6 language settings.}
    \label{fig:ael_case}
\end{figure*}
\begin{figure*}[t]
    \centering
    \includegraphics[width=2.0\columnwidth]{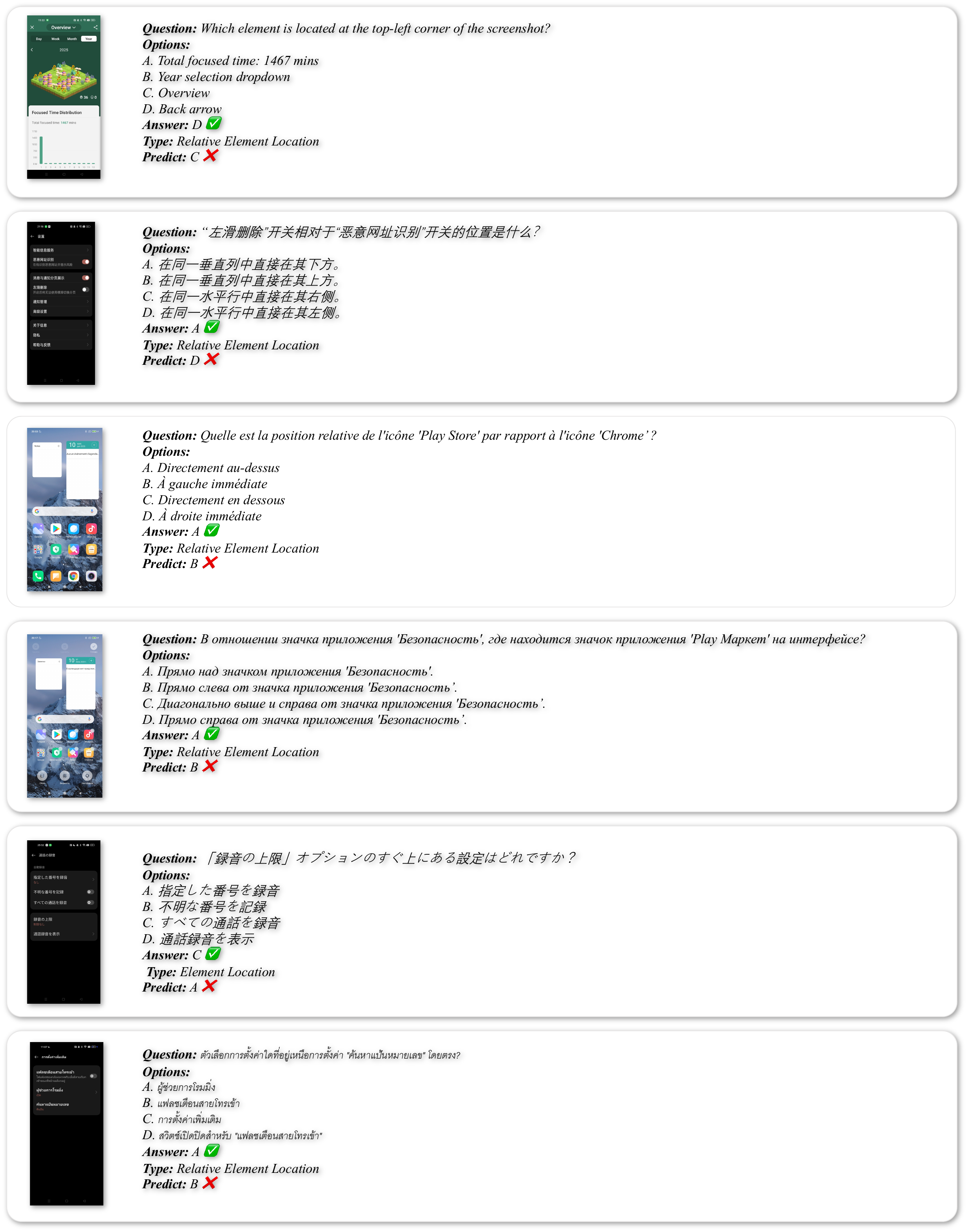}
    \caption{Examples of incorrect responses by LVLMs in REL dimension across 6 language settings.}
    \label{fig:rel_case}
\end{figure*}

\begin{figure*}[t]
    \centering
    \includegraphics[width=2.0\columnwidth]{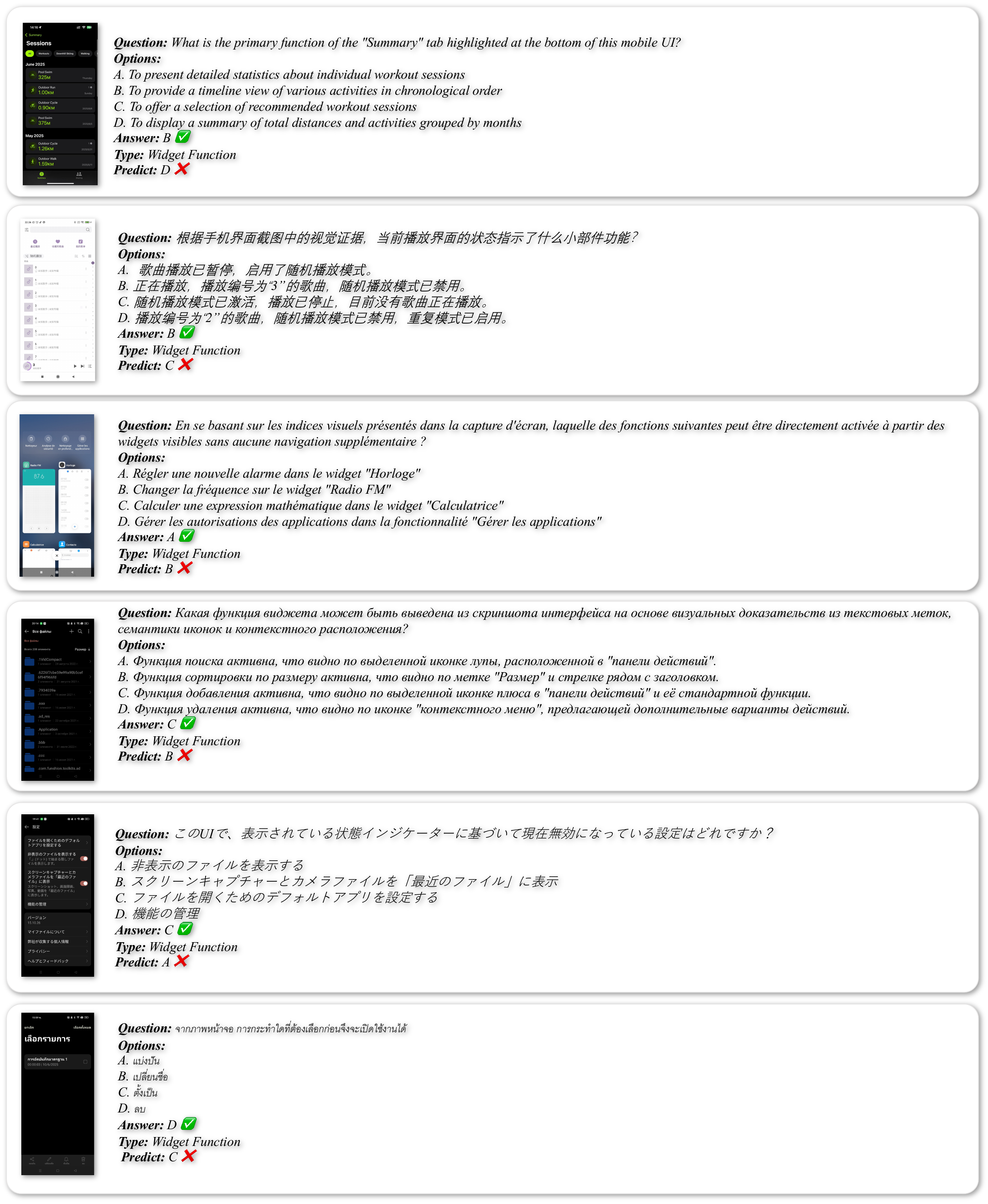}
    \caption{Examples of incorrect responses by LVLMs in WF dimension across 6 language settings.}
    \label{fig:wf_case}
\end{figure*}
\begin{figure*}[t]
    \centering
    \includegraphics[width=2.0\columnwidth]{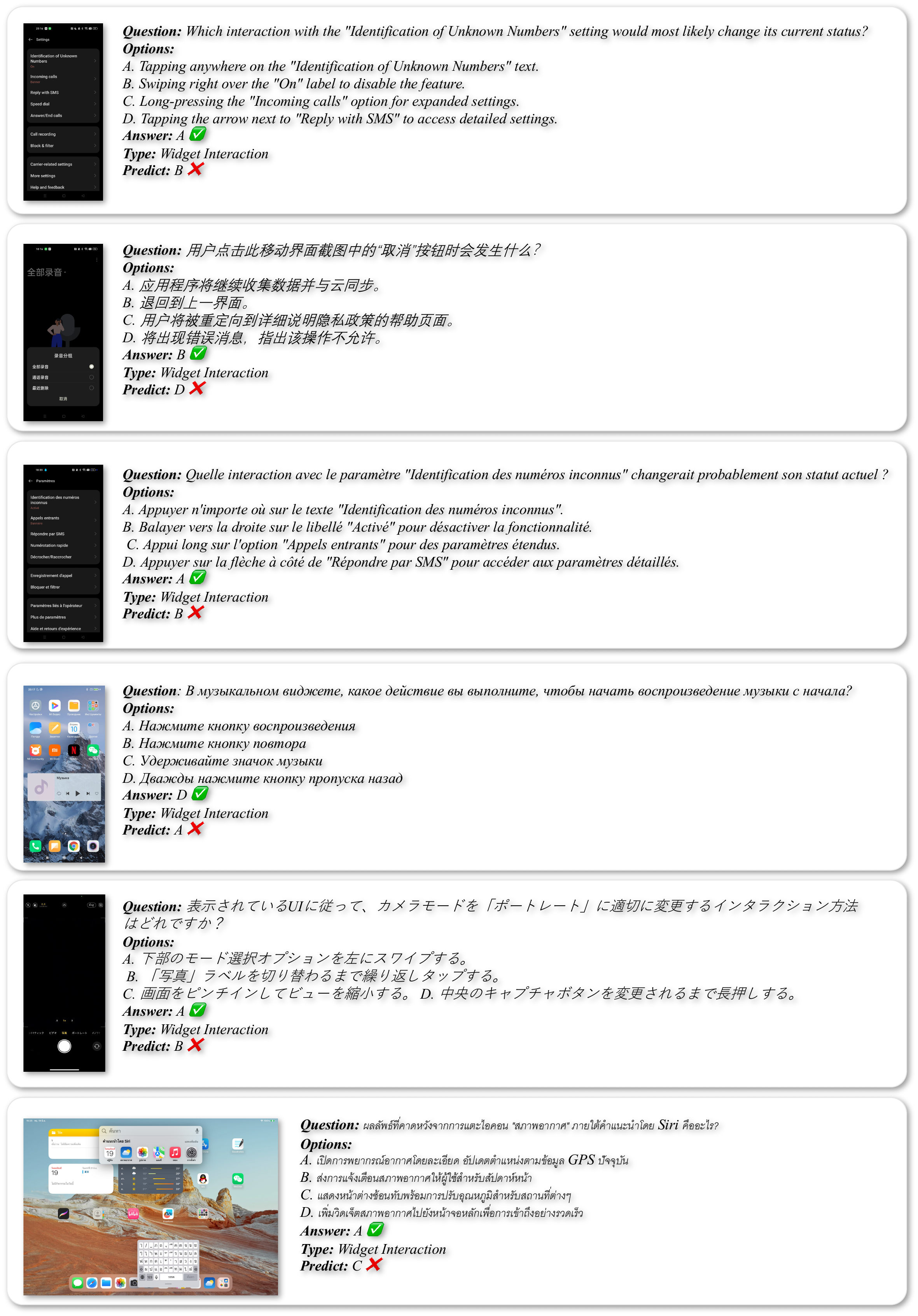}
    \caption{Examples of incorrect responses by LVLMs in WI dimension across 6 language settings.}
    \label{fig:wi_case}
\end{figure*}
\begin{figure*}[t]
    \centering
    \includegraphics[width=2.0\columnwidth]{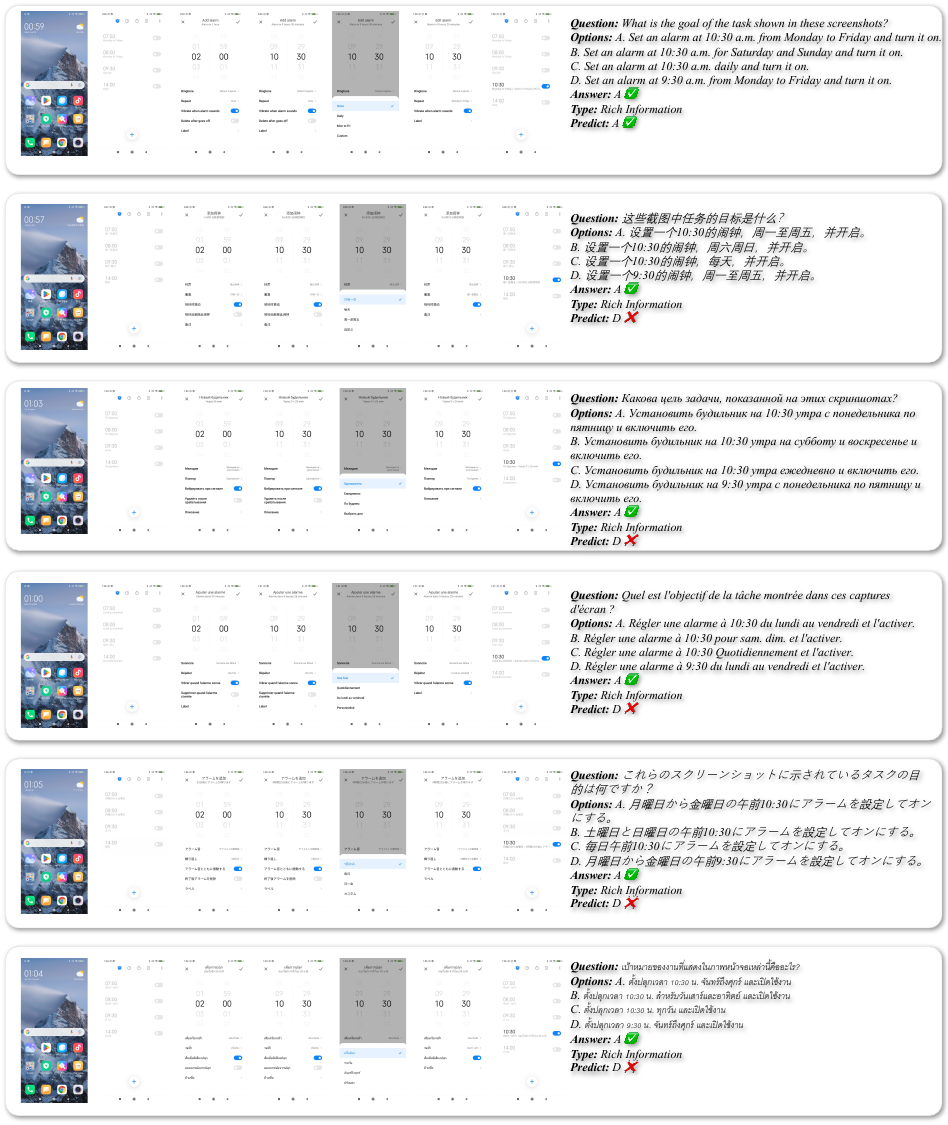}
    \caption{Examples of responses by LVLMs in RI dimension across 6 language settings.}
    \label{fig:ri_case}
\end{figure*}
\begin{figure*}[t]
    \centering
    \includegraphics[width=0.85\textwidth]{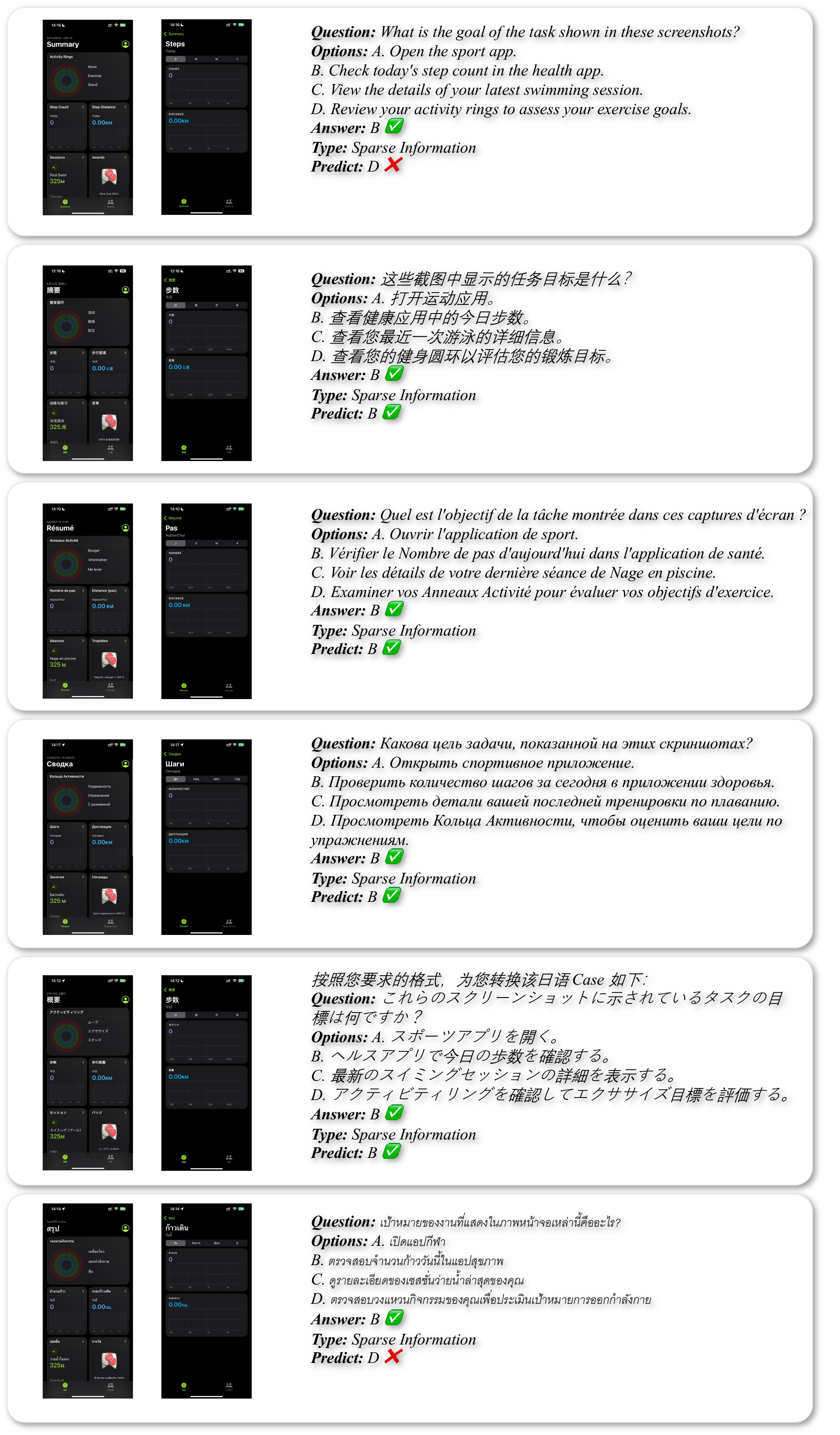}
    \caption{Examples of responses by LVLMs in SI dimension across 6 language settings.}
    \label{fig:si_case}
\end{figure*}

\subsection{Evaluations results on additional models with broader size}
\label{appendix:Evaluations results on additional models with broader size}
As shown in Table~\ref{tab:evaluation on additional models}, to verify the diagnostic power of MPR-GUI-Bench, we extend our evaluation to a broader spectrum of models, including an LVLM for GUI agent: CogAgent-9B~\citep{hong2023cogagent} and two additional open-source baselines: MiniCPM-o 2.6~\citep{yao2024minicpm} and Keye-VL-7B~\citep{kwaikeyeteam2025kwaikeyevl15technical,kwaikeyeteam2025kwaikeyevltechnicalreport}. 

\newcolumntype{C}[1]{>{\centering\arraybackslash}p{#1}}
\newcolumntype{L}[1]{>{\raggedright\arraybackslash}p{#1}}
\begin{table*}[!ht]
\centering
\small
\renewcommand{\arraystretch}{0.9} 
\setlength{\tabcolsep}{5pt} 

\begin{tabular}{ L{3.2cm} C{0.6cm} C{0.85cm} C{0.85cm} C{0.85cm} C{0.85cm} C{0.85cm} C{0.85cm} C{0.85cm} C{0.85cm} C{1.25cm} } 
\toprule 
\multirow{2}{*}{\textbf{Model}} & \multirow{2}{*}{\textbf{Lang}} & \multicolumn{4}{c}{\textbf{Perception}} & \multicolumn{4}{c}{\textbf{Reasoning}} & \multirow{2}{*}{\textbf{$\text{FPR-ACC}$}} \\ 
\cmidrule(lr){3-6} \cmidrule(lr){7-10}
 &  & \textbf{AU} & \textbf{AP} & \textbf{WF} & \textbf{WI} & \textbf{AEL} & \textbf{REL} & \textbf{RI} & \textbf{SI} &  \\ 
\midrule 

\rowcolor[gray]{0.95} \multicolumn{11}{c}{\textit{Open-source LVLMs}} \\

\multirow{6}{*}{MiniCPM-o 2.6} & EN & \cellcolor{CustomGreen!60}\textbf{83.9} & \cellcolor{CustomGreen!60}\textbf{83.6} & \cellcolor{CustomGreen!60}\textbf{81.6} & \cellcolor{CustomGreen!60}\textbf{91.0} & \cellcolor{CustomGreen!60}\textbf{77.9} & \cellcolor{CustomGreen!60}\textbf{84.4} & \cellcolor{CustomGreen!60}\textbf{84.0} & \cellcolor{CustomGreen!60}\textbf{64.0} & \cellcolor{CustomGreen!60}\textbf{79.6} \\ 
& ZH & 76.5 & 82.0 & 75.1 & 90.2 & 75.1 & 73.8 & 80.0 & \cellcolor{CustomGreen!60}\textbf{64.0} & 75.9 \\ 
& FR & 76.5 & 79.0 & 77.8 & 89.9 & 74.3 & 76.8 & 76.0 & 60.0 & 74.6 \\ 
& RU & 74.6 & 79.5 & 72.3 & 86.6 & 72.4 & 73.8 & 64.0 & 56.0 & 70.2 \\ 
& JA & 72.7 & 77.6 & 74.0 & 85.8 & 74.0 & 67.2 & 68.0 & \cellcolor{CustomGreen!60}\textbf{64.0} & 71.7 \\ 
& TH & \cellcolor{CustomOrange!60}66.4 & \cellcolor{CustomOrange!60}74.0 & \cellcolor{CustomOrange!60}60.6 & \cellcolor{CustomOrange!60}78.1 & \cellcolor{CustomOrange!60}63.1 & \cellcolor{CustomOrange!60}42.6 & \cellcolor{CustomOrange!60}52.0 & \cellcolor{CustomOrange!60}40.0 & \cellcolor{CustomOrange!60}57.1 \\ 
\midrule 

\multirow{6}{*}{Keye-VL-7B} & EN & \cellcolor{CustomGreen!60}\textbf{88.3} & \cellcolor{CustomGreen!60}\textbf{83.9} & \cellcolor{CustomGreen!60}\textbf{81.6} & \cellcolor{CustomGreen!60}\textbf{93.7} & \cellcolor{CustomGreen!60}\textbf{79.0} & \cellcolor{CustomGreen!60}\textbf{73.8} & \cellcolor{CustomGreen!60}\textbf{48.0} & \cellcolor{CustomGreen!60}\textbf{64.0} & \cellcolor{CustomGreen!60}\textbf{73.7} \\ 
& ZH & 82.0 & 81.4 & 73.4 & 89.3 & 77.3 & 68.0 & 40.0 & 52.0 & 66.9 \\ 
& FR & 84.2 & 82.5 & 76.4 & 91.0 & 72.4 & 71.6 & 44.0 & 44.0 & 66.5 \\ 
& RU & 80.1 & 82.0 & 72.1 & 86.6 & 75.7 & 68.3 & 44.0 & \cellcolor{CustomOrange!60}28.0 & 61.8 \\ 
& JA & 78.1 & 82.2 & 71.2 & 86.6 & 72.7 & 58.2 & \cellcolor{CustomOrange!60}36.0 & 40.0 & 61.4 \\ 
& TH & \cellcolor{CustomOrange!60}75.1 & \cellcolor{CustomOrange!60}77.3 & \cellcolor{CustomOrange!60}66.0 & \cellcolor{CustomOrange!60}84.4 & \cellcolor{CustomOrange!60}68.3 & \cellcolor{CustomOrange!60}44.0 & \cellcolor{CustomOrange!60}36.0 & 36.0 & \cellcolor{CustomOrange!60}57.0 \\ 
\midrule 
\rowcolor[gray]{0.95} \multicolumn{11}{c}{\textit{Multimodal GUI Agents}} \\

\multirow{6}{*}{CogAgent-9B} & EN & \cellcolor{CustomGreen!60}\textbf{63.8} & \cellcolor{CustomGreen!60}\textbf{78.1} & \cellcolor{CustomGreen!60}\textbf{63.8} & \cellcolor{CustomGreen!60}\textbf{81.9} & 52.0 & \cellcolor{CustomGreen!60}\textbf{40.8} & 44.0 & 36.0 & 54.6 \\ 
& ZH & 62.8 & 74.3 & 59.3 & 78.1 & \cellcolor{CustomGreen!60}\textbf{54.0} & 31.6 & \cellcolor{CustomGreen!60}\textbf{60.0} & 36.0 & \cellcolor{CustomGreen!60}\textbf{55.0} \\ 
& FR & 56.9 & \cellcolor{CustomOrange!60}69.7 & 58.9 & 68.0 & 43.2 & 38.2 & 52.0 & 36.0 & 51.0 \\ 
& RU & 55.2 & 72.4 & \cellcolor{CustomOrange!60}52.4 & \cellcolor{CustomOrange!60}67.2 & 43.9 & 31.6 & 56.0 & \cellcolor{CustomGreen!60}\textbf{52.0} & 53.8 \\ 
& JA & \cellcolor{CustomOrange!60}48.3 & 73.5 & 54.8 & 68.0 & 36.0 & \cellcolor{CustomOrange!60}26.3 & \cellcolor{CustomOrange!60}40.0 & 44.0 & 47.9 \\ 
& TH & 50.0 & \cellcolor{CustomOrange!60}69.7 & 54.8 & 68.6 & \cellcolor{CustomOrange!60}32.0 & 31.1 & \cellcolor{CustomOrange!60}40.0 & \cellcolor{CustomOrange!60}20.0 & \cellcolor{CustomOrange!60}42.8 \\ 

\bottomrule
\end{tabular} 
\caption{Performance results (\%) for the \textbf{three additional baseline models} across all six languages. Within each dimension, the \colorbox{CustomGreen!60}{highest} and \colorbox{CustomOrange!60}{lowest} accuracies are highlighted in green and orange, respectively, to demonstrate the performance variance across different model scales.}
\label{tab:evaluation on additional models}
\small
\end{table*}
\section{Details on GUI-XLI}
\subsection{Memory Construction Details}
\label{aappendix:Memory Construction Details} 
To ensure maximum inference efficiency and real-time responsiveness, we maintain an extremely compact memory for the foundational P\&R capabilities, where each dimension (AU, AP, WF, WI, AEL, and REL) consists of only 10 entries. 

For the reasoning-intensive dimensions (RI and SI), relying solely on their own training samples might be insufficient as they inherently depend on foundational perception capabilities. Therefore, we construct the memory for RI and SI by aggregating entries from the six foundational dimensions: AU, AP, WF, WI, AEL, and REL. This aggregation is performed independently at each layer to maintain the layer-wise alignment structure defined in Section~\ref{sec:GUI-XL-Memory}. This strategy ensures that the intervention for reasoning tasks is grounded in robust fine-grained perception cues.
\subsection{Case before and after using GUI-XLI}
\begin{figure*}[t]
    \centering
    \includegraphics[width=2.0\columnwidth]{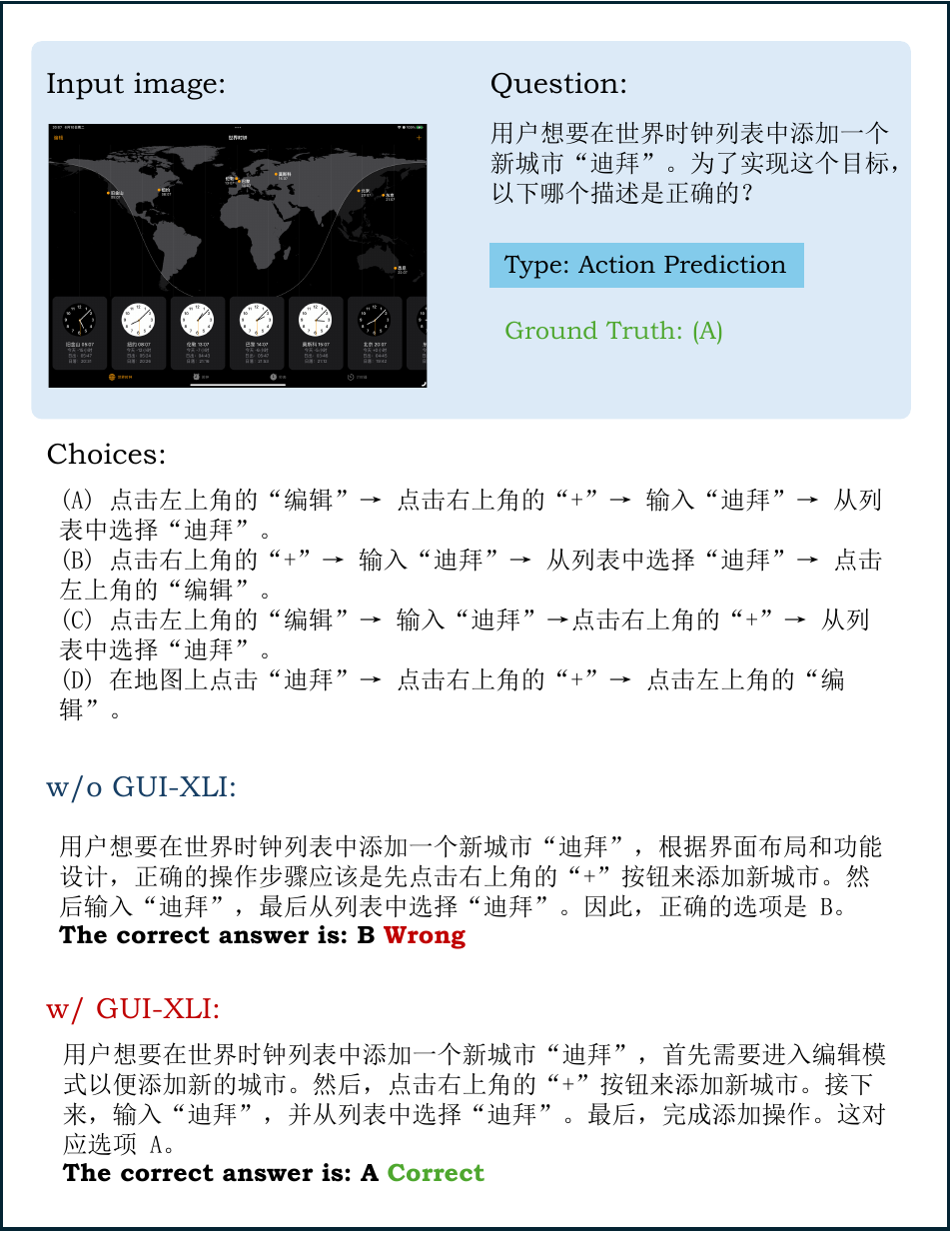}
    \caption{A typical case to show the effectiveness of GUI-XLI.}
    \label{fig:case_gui_xli}
\end{figure*}

As shown in Figure~\ref{fig:case_gui_xli}, we present a typical case where GUI-XLI corrects a failed reasoning process in a Chinese sample, leading to a successful outcome.
\subsection{Overview of GUI-XLI}
\begin{figure*}[t]
    \centering
    \includegraphics[width=2.0\columnwidth]{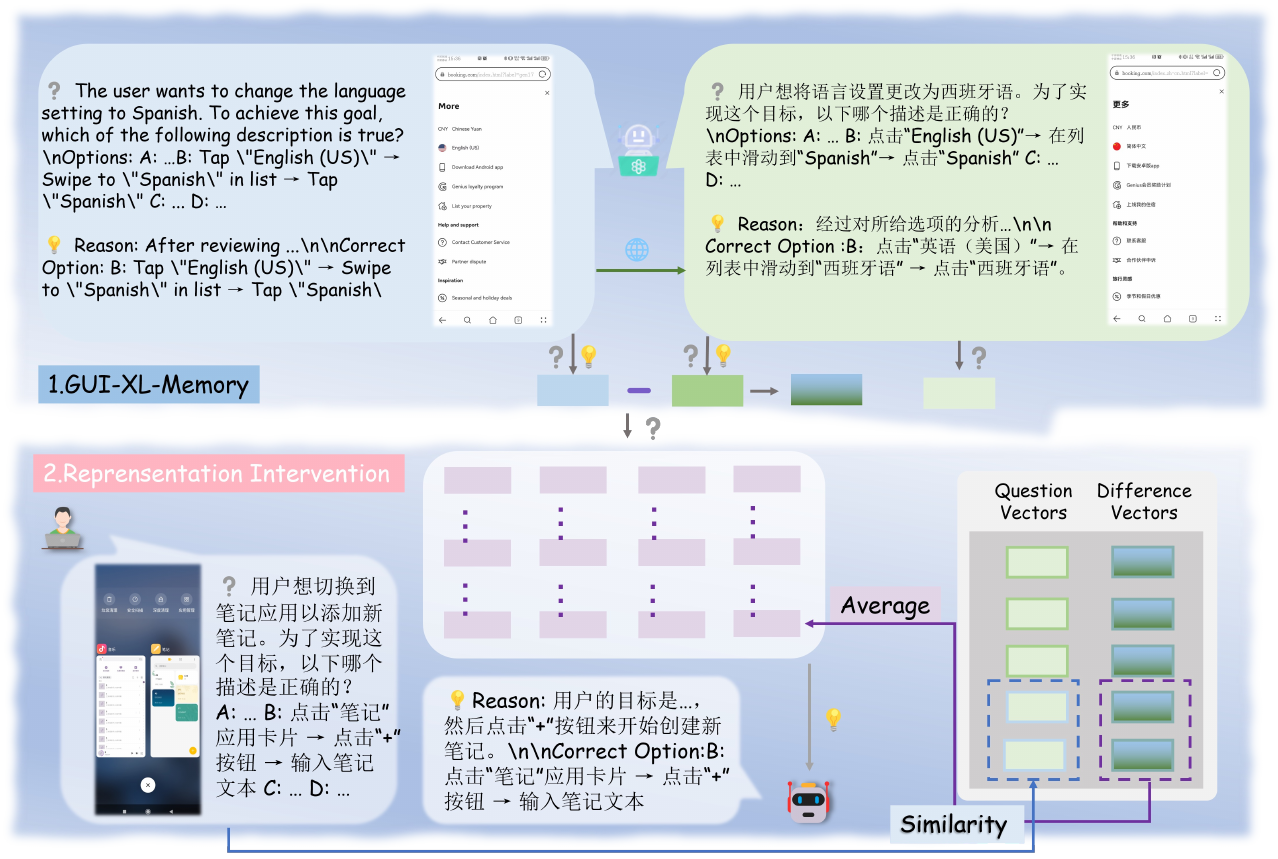}
    \caption{An Overview of our \textbf{GUI-XLI} method.
    \textbf{Step 1 GUI-XL-Memory}: We sample semantically parallel VQA pairs to form entries in \textbf{GUI-XL-Memory}. GUI-XLI uses semantically parallel entries from Step 1 as reference anchors to steer non-English states in Step 2.
    \textbf{Step 2 Cross-lingual Representation Intervention}: When answering non-English questions, related entries are retrieved to calculate cross-lingual discrepancy vectors and then injected to certain layer as intervention to add P\&R capabilities to non-English settings. }
    \label{fig:GUIXLI}

\end{figure*}

In this subsection, we present the overview illustration figure of our GUI-XLI method in Figure~\ref{fig:GUIXLI}.
\label{appendix:Overview of GUI-XLI}

\section{Usage of LLMs}
GPT-4o~\citep{openai2024gpt4ocard} was employed to assist in refining the language and enhancing the readability of the manuscript. All ideas, experiments, analyses, and conclusions were conceived, conducted, and verified by the authors.
\end{document}